\documentclass[10pt,twocolumn,letterpaper]{article}

\usepackage{cvpr}              %
\usepackage{csquotes}
\usepackage{mdframed}
\usepackage{tcolorbox}

\usepackage{float}

\definecolor{cvprblue}{rgb}{0.21,0.49,0.74}

\usepackage[pagebackref,breaklinks,colorlinks,allcolors=cvprblue]{hyperref}

\usepackage{array}
\usepackage{booktabs}
\usepackage{ragged2e}

\def\paperID{2549} %
\def\confName{CVPR}
\def\confYear{2025}

\title{
WordCraft: Interactive Artistic Typography with Attention \\ Awareness and Noise Blending
}

\author{
Zhe Wang$^{1}$, \quad Jingbo Zhang$^{2}$, \quad Tianyi Wei$^{3}$, \quad Wanchao Su$^{4}$, \quad Can Wang$^{5*}$ \\ \vspace{-0.5mm}
\footnotesize $^{1}$Jiangxi University of Finance and Economics\quad
\footnotesize $^{2}$Tencent Robotics X Lab\\ \vspace{-0.5mm}  
\footnotesize $^{3}$Nanyang Technological University \quad
\footnotesize $^{4}$Monash University \quad
\footnotesize $^{5}$Hong Kong University\\
\footnotesize $^*$Corresponding author
}

\begin{document}

\twocolumn[{%
\renewcommand\twocolumn[1][]{#1}%
\maketitle
\vspace{-7mm}

\includegraphics[width=1.0\textwidth]{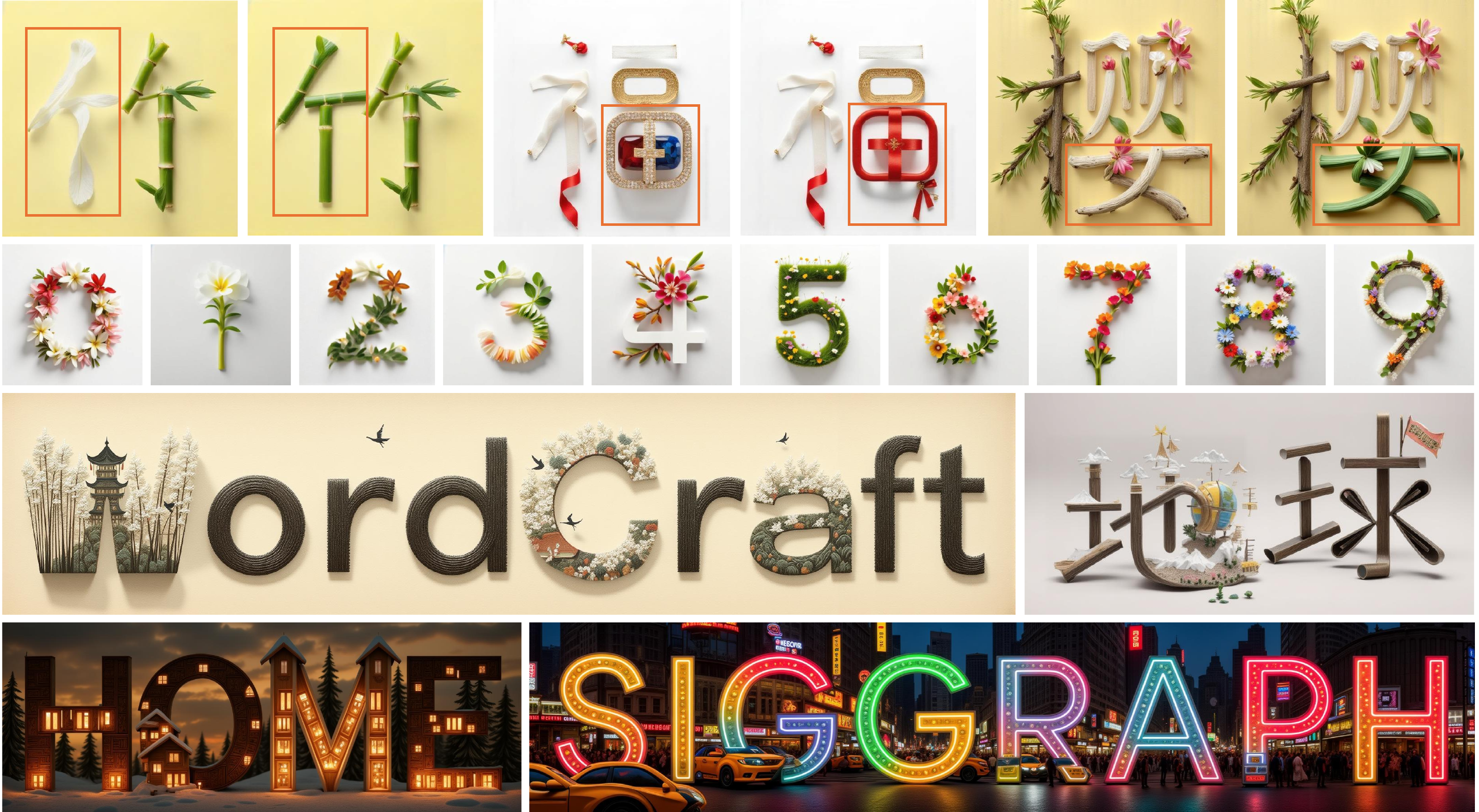}
\vspace{-5mm}
\captionof{figure}{\textit{WordCraft} introduces an interactive text-driven artistic typography system that provides fine-grained control and high aesthetic quality for both single- and multi-character generation across multiple languages. The first row demonstrates the continuous regional editing process, with modified radicals highlighted in orange rectangles. The third row showcases the results of multi-regional editing, such as the characters `W', `c', and `a' in the word `WordCraft', while the remaining characters are generated globally.}
\vspace{0.5em}
\label{fig:teaser}
}]

\begin{abstract}
  Artistic typography aims to stylize input characters with visual effects that are both creative
  and legible. Traditional approaches rely heavily on manual design, while recent generative models, particularly diffusion-based methods, have enabled automated character stylization.
  However, existing
  solutions remain limited in interactivity, lacking support for localized edits, iterative refinement, multi-character composition, and open-ended prompt interpretation.
  We introduce \textit{WordCraft}, an interactive artistic typography system that integrates diffusion models to address these limitations. \textit{WordCraft} features a training-free regional attention mechanism for precise, multi-region generation and a noise blending that supports
  continuous refinement without compromising
  visual quality. To support flexible, intent-driven generation, 
  we incorporate
  a large language model to parse and structure both concrete and abstract user prompts.
  These components allow our framework
  to synthesize high-quality, stylized
  typography across single- and multi-character inputs across multiple languages, supporting diverse user-centered workflows.
  Our system significantly enhances
  interactivity in artistic typography synthesis, opening up creative possibilities for artists and designers.
\end{abstract}
\section{Introduction}

Artistic typography aims to transfer visual effects onto target text in raster or vector form. 
It has found wide applications in commercial advertising, artistic design, and education.
Traditional artistic typography depends on manual craftsmanship and artistic expertise to achieve the desired visual effects, making it time-consuming, labor-intensive, and impractical for large-scale application.

Recent advances in generative models, such as GANs and diffusion models, have facilitated
the artistic typography process by framing it as neural style transfer \cite{jing2019neural,fu2023neural,azadi2018multi,mao2022intelligent,wang2019typography}, image-to-image translation \cite{isola2017image,zhu2017unpaired,yang2019tet}, or conditional generation \cite{hemetadesigner,he2023wordart,tanveer2023ds,wang2023anything,feng2024vitaglyph}. In particular, text-conditional diffusion models \cite{rombach2022high,zhang2023adding,tan2024ominicontrol} have significantly improved productivity and visual diversity of stylized text generation by allowing users to specify visual effects through text prompts.
However, generating
artistic typography in an interactive manner remains challenging due to several limitations: 1) \textbf{Local stylization.} Users often need to stylize specific 
characters like Chinese—while existing methods primarily focus on global styling; 2) \textbf{Continuous refinement. }Users may wish to iteratively edit or refine specific regions of the generated typography without altering the rest, but current models operate in a single-step manner and lack support for continuous refinement; 3) \textbf{Multi-character support. }Input source image may include single or multi-character combinations, yet most current methods only support single-character generation. When a single prompt describes different styles for each character in a combination, the scenario becomes more complex—an issue not addressed in existing works; 4) \textbf{Open-set text guidance. }User-provided prompts can be arbitrary and unconstrained, ranging from concrete to abstract, and may describe global styles or localized effects. However, existing methods are limited to handling single, straightforward, and concrete text prompts.
These challenges significantly limit user interactivity and hinder the overall user experience.

To address these challenges, we propose \textit{WordCraft}, an interactive artistic typography system guided by user prompts and powered by recent diffusion models. 
We introduce a regional attention mechanism within the diffusion process to enable multi-region editing, as well as a noise blending technique during denoising to support continuous edits. These techniques are designed to handle both single-character and multi-character inputs. Additionally, we employ
a large language model (LLM) to interpret user intent and structure prompts, enabling flexible and open-ended text descriptions. As shown in Fig.~\ref{fig:teaser}, our method
generates visually appealing results for both single and multi-character inputs across multiple languages, even when handling complex or abstract text prompts.

In summary, our contributions
are threefold:
\begin{itemize}
    \item \textbf{A regional attention for multi-region
    generation.} \textit{WordCraft} introduces a novel training-free regional attention mechanism within the diffusion model, enabling precise and independent control over different regions of a typographic image. This supports flexible layout and style generation across both single-character and multi-character inputs.
    
    \item \textbf{Noise blending for continuous editing.} To support coherent iterative refinement, 
    we propose
    a noise blending technique during the diffusion denoising process, allowing smooth and consistent updates without sacrificing image quality or structure.
    This technique also supports both single-character and multi-character inputs.
    
    \item \textbf{User intent-aware typography generation.} \textit{WordCraft} integrates a LLM to interpret user intent and structure text prompts, enabling intuitive and open-ended interactions. This facilitates the generation of visually compelling artistic typography, even from abstract or complex descriptions.

\end{itemize}

\section{Related Work}

\noindent\textbf{Artistic Text Synthesis.} 
Recent advances in GAN have enabled impressive progress in artistic text synthesis \cite{yang2019tet,yang2019controllable,yang2021shape,mao2022intelligent}. For instance, \textit{TET-GAN} \cite{yang2019tet} disentangles and recombines content and style features via deep neural representations, supporting high-quality text effects transfer and one-shot style learning. To further improve texture control, \textit{ShapeMatching GAN} \cite{yang2019controllable} enables adjustable glyph deformation using bidirectional shape matching and multi-scale feature transfer.
More recently, diffusion models have further advanced this field, enabling the generation of highly detailed and visually compelling artistic text images \cite{wang2023anything,tanveer2023ds,he2023wordart,iluz2023word,hemetadesigner}. For example, \textit{MetaDesigner} \cite{hemetadesigner} leverages differentiable rasterization and a depth-to-image stable diffusion model to iteratively optimize the vector parameters of an SVG representation for semantic glyph transformation. It then employs \textit{ControlNet} \cite{zhang2023adding} to enrich texture styles guided by edges, depth maps, or scribbles. 
\textit{VitaGlyph} \cite{feng2024vitaglyph} divides the input glyph image into subject and surrounding regions, then employs two independent ControlNets to synthesize and fuse these regions for artistic text generation.
However, these methods ignore the aforementioned challenges, which our method tackles, thus restricting user interaction and control.
It should be noted that current methods typically focus on two output types: raster images and vector graphics. While vector graphics offer scalability and resolution independence, raster images provide richer textures and finer visual effects. To better capture expressive styles and detailed textures, our method—like \textit{MetaDesigner} and \textit{VitaGlyph}—chooses raster images as the synthesis output.

\begin{figure*}
  \centering
  \includegraphics[width=0.96\textwidth]{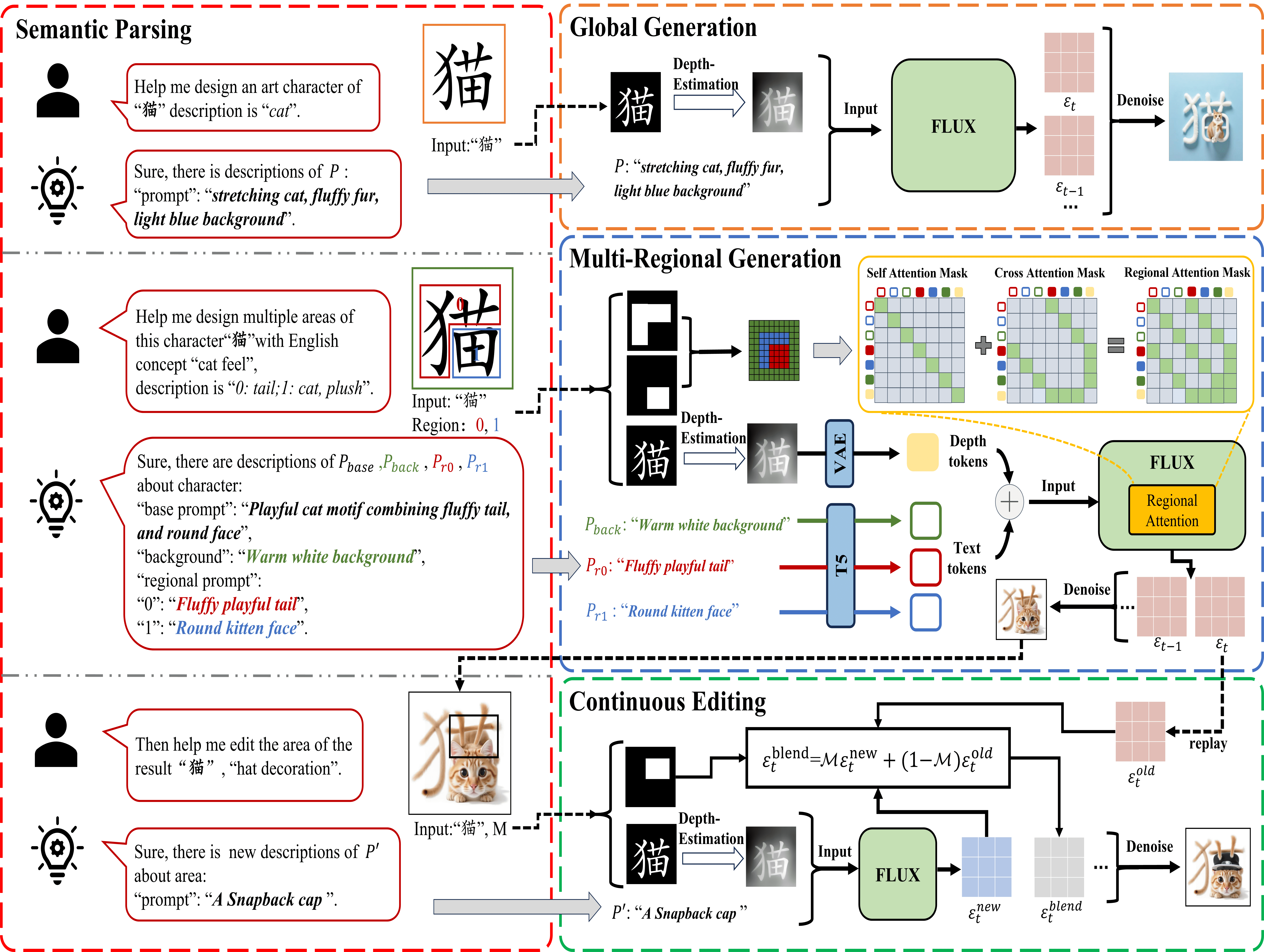}
  \caption{\textbf{Framework. }Our method supports stylized generation, regional editing, and multi-step user interaction.}
  \label{fig:framework}
\end{figure*}

\noindent\textbf{Text-to-Image Synthesis.} 
Significant advances in diffusion models have greatly improved text-to-image synthesis. To reduce training and inference costs while enhancing generation quality, many approaches compress images into latent representations and perform diffusion in this space, as demonstrated by models such as \textit{Imagen} \cite{saharia2022photorealistic}, \textit{DALLE-2} \cite{ramesh2022hierarchical}, and \textit{LDM} \cite{rombach2022high}.
To enable conditional generation, \textit{ControlNet} \cite{zhang2023adding} was introduced to guide the denoising process using external inputs such as segmentation maps or depth cues, inspiring a series of ControlNet-based extensions \cite{zhao2023uni,zhao2025local,liu2024smartcontrol,bhat2024loosecontrol}. Recently, several studies have explored applying ControlNet to artistic typography synthesis \cite{he2023wordart,hemetadesigner,
mu2024fontstudio,yang2023glyphcontrol}. 
Our method also builds upon the ControlNet framework; however, unlike prior approaches, it introduces regional attention and noise blending techniques into the diffusion model, enabling a more user-friendly and interactive experience for artistic typography generation.

\noindent\textbf{Large Language Models for User Understanding.} 
Large Language Models (LLMs) have made continuous progress and are now widely applied across diverse domains \cite{yao2024survey,zheng2023judging,nam2024using,ahn2024large}. In particular, they have proven effective for task understanding and planning. For example, \textit{WordArt Designer} \cite{he2023wordart} parses user descriptions of desired text styles into fixed prompt templates, but it still requires users to follow a rigid and predefined input format. In contrast, \textit{Chat2Layout} \cite{wang2024chat2layout} defines a set of atomic operations for 3D layout synthesis and leverages LLMs to decompose open-ended instructions into a sequence of these atomic tasks, thereby supporting a broad and flexible vocabulary.
Inspired by these approaches, we categorize diverse user intentions expressed through text prompts in artistic typography and employ an LLM to convert open-set inputs into structured representations for our model. Unlike previous methods that rely on fixed and narrowly scoped prompts, our approach enables flexible interpretation of user intent, substantially enhancing interactivity and usability in artistic typography generation.
\section{Method}

Our \textit{WordCraft} system facilitates a user-centered, multi-region controllable, and iteratively optimizable editing workflow by integrating a suite of editing modules powered by a LLM, such as GPT-4. As depicted in Fig.~\ref{fig:framework}, \textit{WordCraft} comprises three core modules: (1) \textbf{Semantic Parsing}, (2) \textbf{Multi-Regional Generation}, and (3) \textbf{Continuous Editing}. The semantic parsing module interprets user intent and decomposes semantic layout tasks, while also parameterizing the input character into a glyph image. The multi-regional generation module enables fine-grained, customized creation and modification of different glyph regions. Finally, the continuous editing module supports iterative, region-specific refinement of the existing artistic typography image.

\subsection{Semantic Parsing}

\subsubsection{LLM-Based Task Decomposition} 

\begin{figure}
    \centering
    \includegraphics[width=0.99\linewidth]{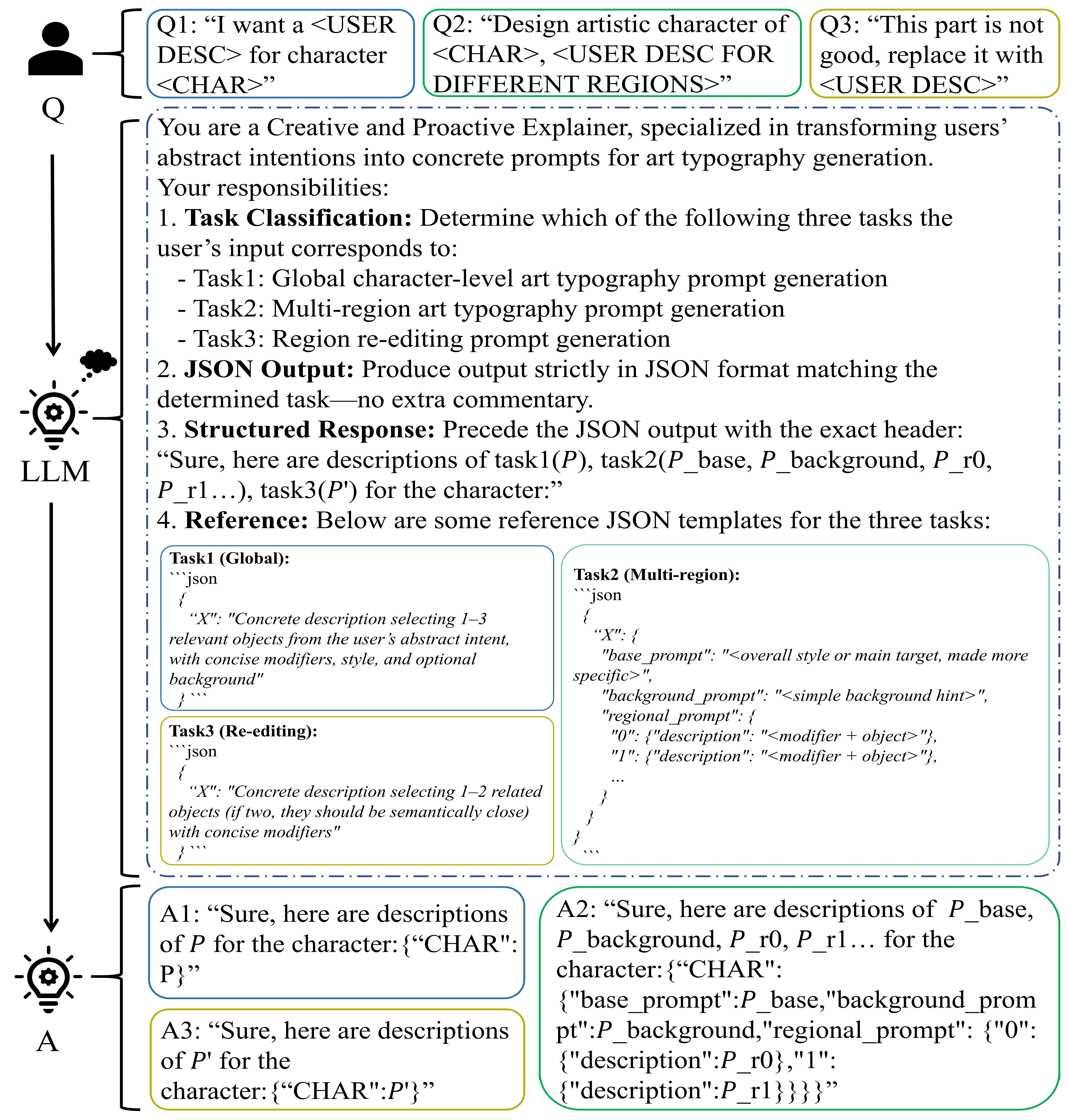}
    \caption{\textbf{LLM for Task Decomposition.} For Q2 and Q3, users could also provide the regional masks.}
    \label{fig:prompt-parsing}
\end{figure}

The semantic parsing module serves as a critical component in our system, responsible for converting open-ended user descriptions into structured semantic representations tailored to a variety of generation or editing tasks. Specifically, we define three task types based on the potential requirements of artistic font processing: global generation, multi-regional generation, and continuous editing. As shown in Fig.~\ref{fig:prompt-parsing}, we first employ GPT-4 as a knowledge engine to deeply understand users’ task-specific intentions and automatically generate semantic prompts adapted to the three editing scenarios described above. This approach lowers the operational barrier for users and minimizes perceptual differences. In essence, users only need to provide a general description $Q$ of the desired font stylization and specify the intended editing regions $I_{region}$ (if applicable), while our semantic parsing module handles these inputs and output the parsed prompts $P$ and the target character $c$ with associated masks:
\begin{equation}
P, c = \text{GPT}(Q, I_{region}),
\label{eq:gpt}
\end{equation}
where $P$ is formatted as a JSON document, with different structures tailored to each specific task. For example, in the global generation task, the parsed $P$ contains only a single global base comprehensive prompt $P_b$. In contrast, for the multi-regional generation task, $P$ includes not only the base prompt $P_b$ but also additional regional prompts $[P_1, \cdots, P_N]$, where $N$ indicates the total number of regions. For the continuous editing task, the parsed $P$ specifies prompts $[P_1, \cdots, P_N]$ corresponding to the designated continuous editing regions.

\subsubsection{Character Parameterization}
After obtaining the parsed target character $c$, we need to further convert it into a glyph image $I$ to ensure that user-selected regions correspond precisely to specific areas within the glyph. To accomplish this, we first extract the relevant font contour from the FreeType library \cite{turner1996freetype} and approximate it with cubic Bézier curves via spline fitting. These curves are then rasterized into a high-resolution, differentiable character image $I_c$ using Differentiable Vector Graphics (DiffVG) \cite{li2020differentiable}, thereby enabling precise region mapping for subsequent generation and editing stages.

\subsection{Multi-Regional Generation}
To generate stylized fonts that meet user expectations, we propose a multi-regional generation technique based on the FLUX.1 model \cite{flux2024}, conditioned on both the parsed prompts and the parameterized character. To ensure that the fundamental structure of the generated stylized font remains consistent with the input character, we use a pre-trained monocular depth estimation model \cite{depthanything} to infer a depth map $I_d$ from the parameterized character image $I_c$. This depth $I_d$, together with the parsed prompts $P$, serves as a condition to guide the generation of the desired stylized font.

\subsubsection{Global Generation}
In order to inject text and depth conditions into the generation pipeline of the FLUX model, we first use the T5 text encoder \cite{podell2023sdxl} and the Variational Autoencoder (VAE) module, both integrated within the FLUX model , to encode the parsed base prompt $P_b$ and the depth map $I_d$ into text tokens $T_b$ and depth tokens $D$, respectively. Both the text and depth tokens are then fed into the FLUX diffusion model and perform multi-model attention (MMA)~\cite{pan2020multi} to guide the stylized font generation. Specifically, the text tokens $T_b$ and depth tokens $D$ are concatenated with the noisy image latent tokens $X$ in the diffusion transformer blocks. Afterwards, the MMA is implemented on the concatenated sequence:
\begin{equation}
\hat{X}, \hat{T_b}, \hat{D} = \text{MMA}([X; T_b; D]),
\label{eq:mma}
\end{equation}
where $\hat{X}$, $\hat{T_b}$, and $\hat{D}$ represent the corresponding tokens that have the same shape as the input token after passing through the MMA blocks.
After training, the FLUX diffusion model enables generating a global stylized font with semantics related to the input prompts and consistent with the source character structure.

\subsubsection{Regional Attention for Multi-regional Generation.}
To perform multi-regional generation, inspired by \cite{zhang2024humanref}, we introduce a multi-region-aware attention mask into the above MMA process. Specifically, we additionally encode the parsed regional prompts $[P_1, \cdots, P_N]$ as regional text tokens $[T_1, \cdots, T_N]$, and concatenate them with other tokens into a sequence $[X; T_b; T_1; \cdots; T_N; D]$. The token sequence is then fed to the attention blocks in Eq.~\ref{eq:mma} to perform multi-region-aware MMA:
\begin{equation}
\text{MMA}([X; T_b; T_1; \cdots; T_N; D], M) = \text{softmax}\left(\frac{qk^\top}{{d_k}} \odot M \right)v,
\label{eq:regional_mma}
\end{equation}
where $q$, $k$, and $v$ indicate the query, key, and value vectors in the transformer attention computation, respectively. $d_k$ is the key dimension. $M$ indicates the regional attention mask, which is defined as follows:
\begin{equation}
M =
\begin{bmatrix}
M_{X2X} & \cdots & M_{X2T_k} & \cdots & M_{X2D} \\
\cdots & & & & \cdots \\
M_{T_k2X} & \cdots & M_{T_k2T_k} & \cdots & M_{T_k2D} \\
\cdots & & & & \cdots \\
M_{D2X} & \cdots & M_{D2T_k} & \cdots & M_{D2D}
\end{bmatrix},
\label{eq:regional_mask}
\end{equation}
where each mask block $M_{\cdot2\cdot}$ indicates the attention relationship between different token types, which is used for governing interaction between different token types. In fact, the mask blocks on the diagonal of the matrix ($M_{X2X}$, $M_{T_k2T_k}$, and $M_{D2D}$) essentially represent the self-attention mask, while the rest mask blocks in the upper and lower triangular areas represent the self-attention mask. 

To prevent information leakage and confusion between regions, semantically related image regions within the self-attention mask block $M_{X2X}$ should pay attention to each other, while the image regions corresponding to the regional token $T_k$ in the cross-attention mask block $M_{X2T_k}$ should similarly interact with one another. Therefore, we define the mask blocks in Eq.~\ref{eq:regional_mask} as follows:

\noindent \textbf{Image-to-Image Mask Block}:
\begin{equation}
M_{X2X}[i, j] =
  \begin{cases}
  1, & \text{if} \ i, j \in \Omega_k \\
  0, & \text{otherwise}
  \end{cases},
\label{eq:i2i}
\end{equation}
where $\Omega_k$ ($k \in [1,\cdots,N]$) denotes the \textit{k}-th regional mask derived from the intended editing regions $I_{region}$ specified by the user. 

\noindent \textbf{Image-to-Text Mask Blocks}:
\begin{equation}
M_{X2T_k}[i, j] = M_{T_k2X}[j, i] = 
    \begin{cases}
    1, & \text{if} \ i \in \Omega_k \\
    0, & \text{otherwise}
    \end{cases},
\label{eq:i2t}
\end{equation}

\noindent \textbf{Image-to-Depth Mask Blocks}:
\begin{equation}
M_{X2D}[i, j] = M_{D2X}[j, i] = 
    \begin{cases}
    1, & \text{if} \ i, j \in \Omega_k \\
    0, & \text{otherwise}
    \end{cases},
\label{eq:i2d}
\end{equation}

\noindent \textbf{Other Mask Blocks}:
\begin{equation}
\begin{split}
M_{T_k2T_l} &= 
\begin{cases}
\mathbf{1},  \quad \text{when} \ k = l \\
\mathbf{0}, \quad \text{when} \ k \neq l
\end{cases} \\
M_{T_k2D} &= M_{D2T_k} = \mathbf{0}, \\
M_{D2D} &= \mathbf{1}.
\end{split}
\label{eq:other_cross}
\end{equation}
where $M_{T_k2T_l}$ ($k \neq l$) and $M_{T_k2D}$ are explicitly set to zero matrices, indicating that there is no attention interaction between different regional prompts and depths. This design ensures that the structural and semantic features of the generated image are guided independently by the depth and the corresponding regional prompts, respectively.

\begin{figure*}
    \centering
    \includegraphics[width=1.0\linewidth]{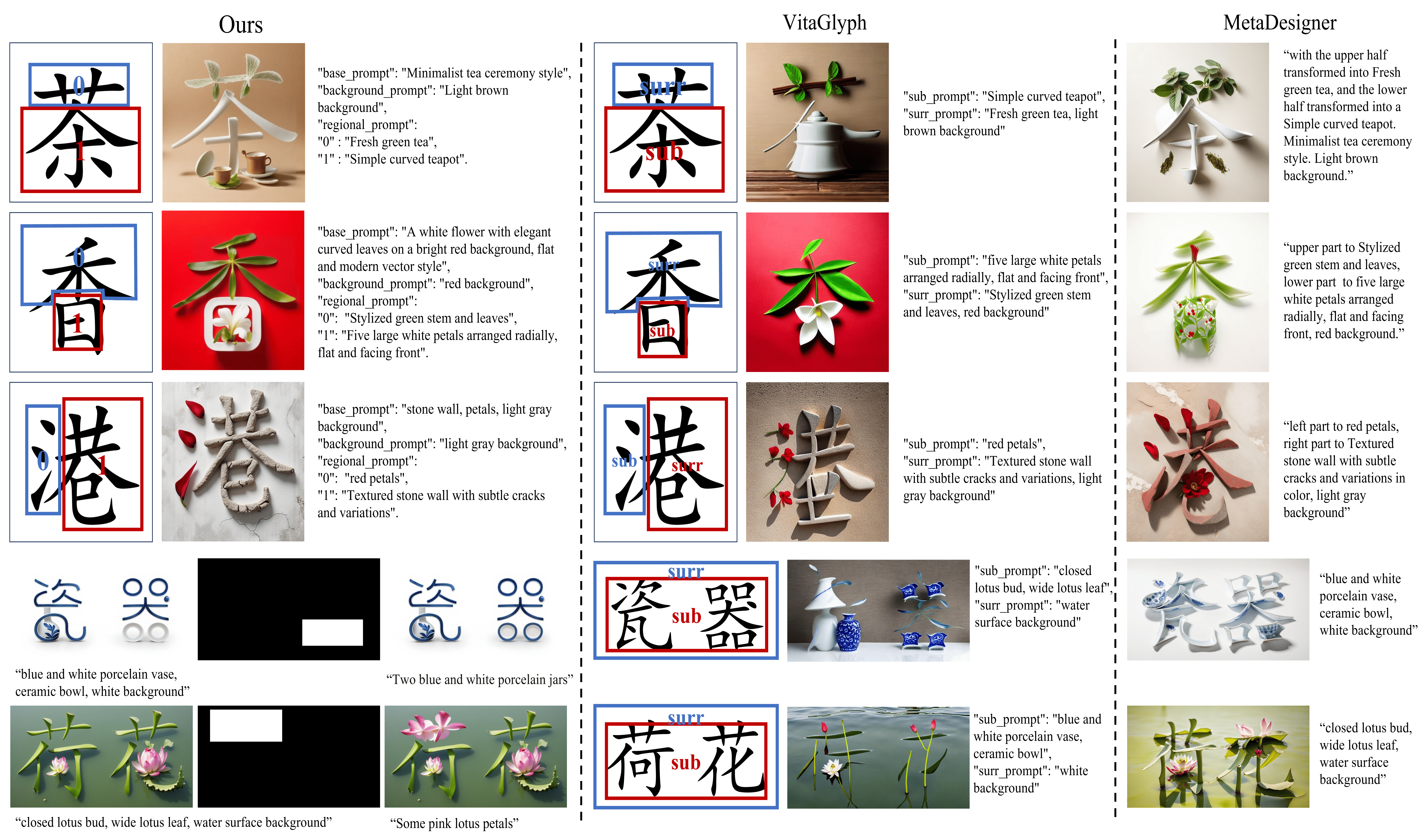}
    \caption{\textbf{Comparison.} We compare our method with \textit{VitaGlyph} and \textit{MetaDesigner}. Unlike \textit{VitaGlyph}, which supports only two regions (the subject and the surrounding area), and \textit{MetaDesigner}, which relies on a single text prompts to describe each region, our method supports multi-regional generation using various masks (see also Fig.~\ref{fig:multiregion}). Additionally, our approach enables continue editing, as demonstrated in the last two multi-character examples—application not supported by either \textit{VitaGlyph} or \textit{MetaDesigner}. Therefore, we do not include continue editing results for \textit{VitaGlyph} or \textit{MetaDesigner} in this comparison.}
    \label{fig:cmp}
\end{figure*}

\subsection{Continuous Editing}

To achieve precise and repeatable local editing while preserving the overall structure of generated glyphs, we introduce a noise blending (NB) strategy. This strategy empowers users to selectively modify specific regions of a given stylized font, while ensuring that unedited regions remain unaffected.

In contrast to conventional inpainting or full-image regeneration techniques, our approach injects newly sampled noise exclusively into user-specified regions, retaining the original noise in all other areas. This selective noise replacement enables an efficient and lightweight editing pipeline, offering fine-grained control over the editing process. Specifically, we first deduce corresponding binary spatial masks $\Omega_k$ ($k \in [1,\cdots,N]$, where $N$ is the number of regions to be edited) from the user-specified editing regions $I_{region}$. For each region, our semantic parsing module produces a text prompt $P_k$ that describes the desired appearance within the masked area. Subsequently, we perform $t$-step denoising on the input image to extract the original noise sequence $\hat{\epsilon}^{old}_t$. Using the FLUX diffusion model, we then conduct forward denoising conditioned on the multi-region text prompts and the depth map. At each denoising step, for each region $k$, a new noise $\hat{\epsilon}^{new}_{t,k}$ is predicted based on the corresponding prompt $P_k$. The final mixed noise $\hat{\epsilon}^{blend}_t$ is computed by blending the new and original noises according to the binary masks:
\begin{equation}
\hat{\epsilon}^{blend}_t = [1 - (\Omega_1 \cup \cdots \cup \Omega_N)] \cdot \hat{\epsilon}^{old}_t + \Omega_1 \cdot \hat{\epsilon}^{new}_{t,1} + \cdots + \Omega_N \cdot \hat{\epsilon}^{new}_{t,N}
\label{eq:bne}
\end{equation}
where each mask $\Omega_k$ is preprocessed to match the spatial resolution of the noise representation. This blending mechanism ensures that only the designated regions are influenced by the new guidance, while the remainder of the image retains its original characteristics. Finally, the blended noise is reintroduced into the diffusion process to update the latent representation. Upon completion of all denoising steps, the final edited image is decoded from the latent space.

Notably, this strategy inherently supports iterative refinement, enabling users to repeatedly adjust the same region or apply distinct cues to different sub-regions. As a result, our NB strategy provides users with fine-grained and flexible control over local modifications in stylized font generation.

\section{Experiments}

\begin{figure*}
    \centering
    \includegraphics[width=1.0\linewidth]{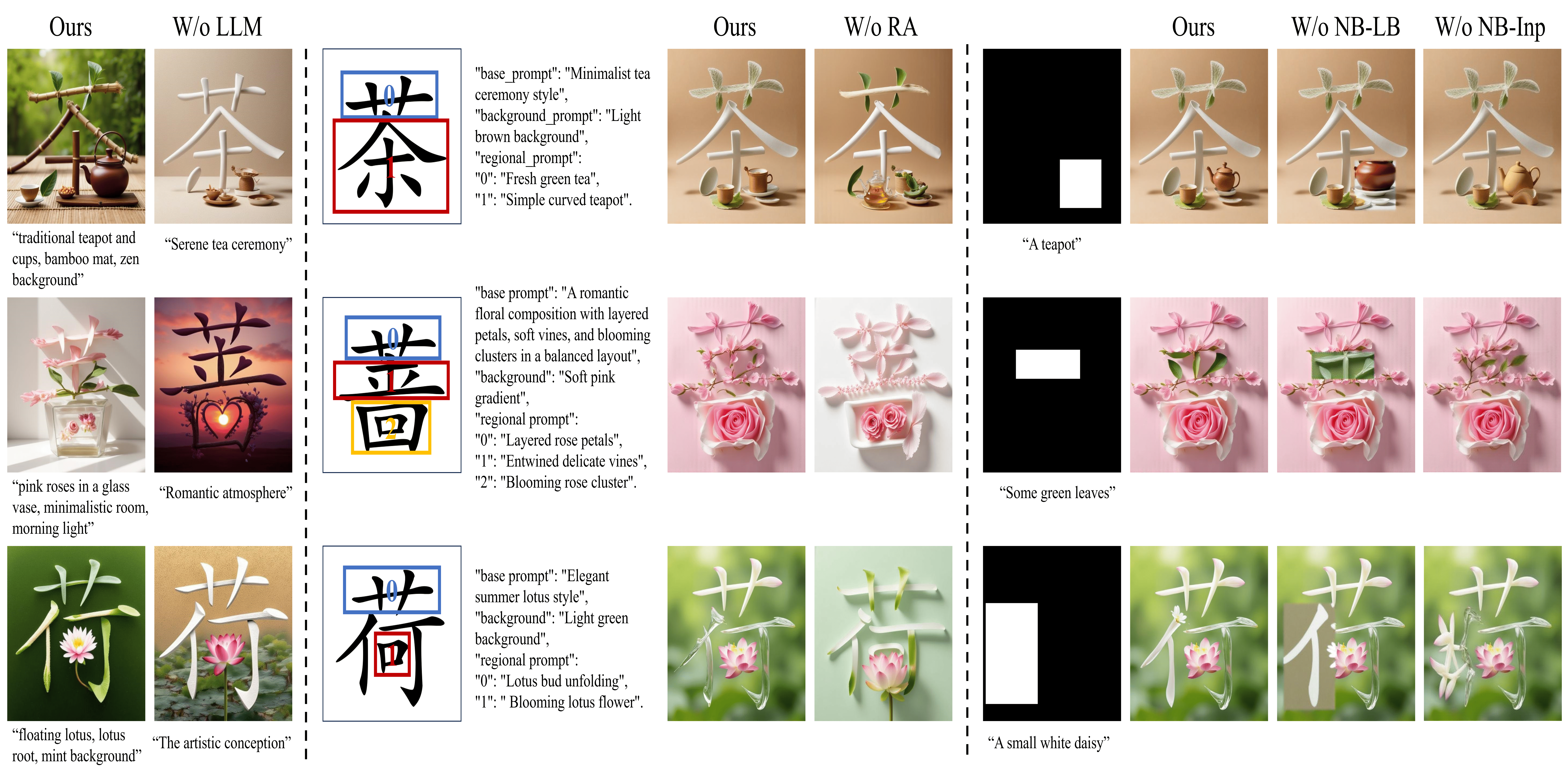}
    \caption{\textbf{Ablation Study.} We compare our full method with three ablated variants: without the LLM component (\textit{w/o LLM}), without rational attention (\textit{w/o RA}), and without noise blending (\textit{w/o NB-LB} and \textit{w/o NB-Inp}).}
    \label{fig:albation}
\end{figure*}

\begin{figure}
    \centering
    \includegraphics[width=0.95\linewidth]{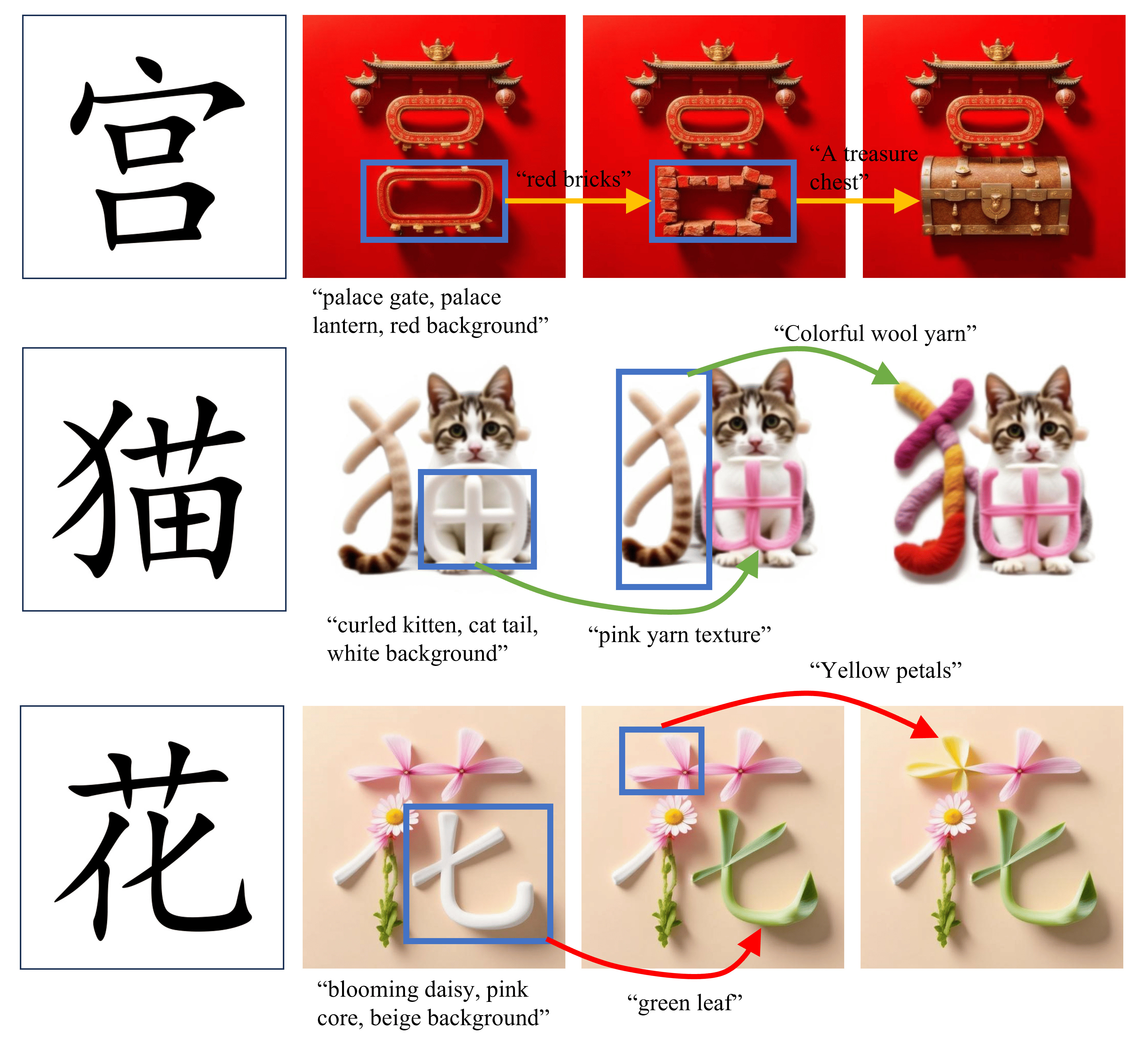}
    \caption{\textbf{Continuous Editing Results.} After the global generation, our method supports continuous user-local editing until the desired results are achieved.}
    \label{fig:continuous}
\end{figure}

\begin{figure}
    \centering
    \includegraphics[width=0.95\linewidth]{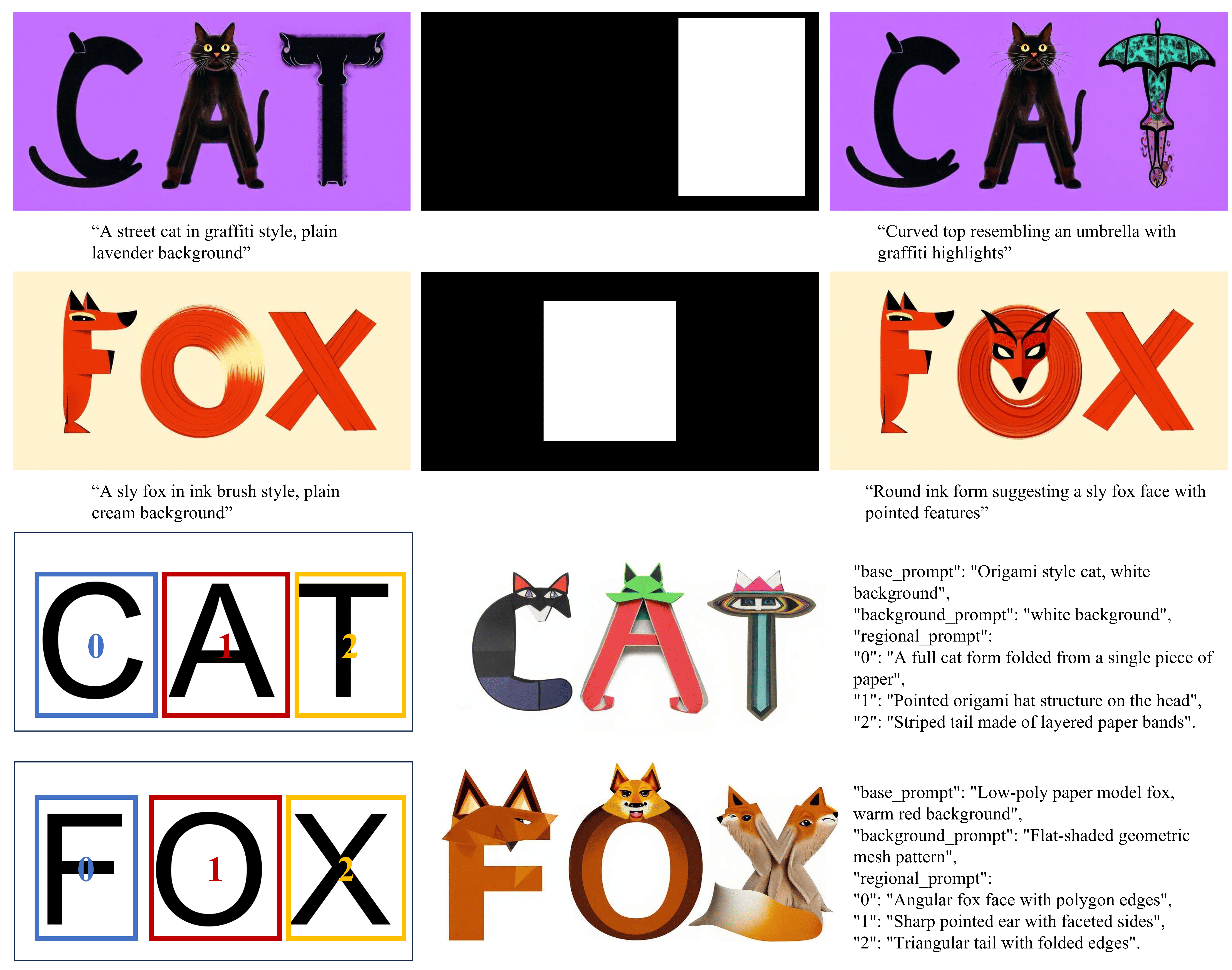}
    \caption{\textbf{Results on English Characters.} Our method also supports multi-regional generation and continuous editing for English characters.}
    \label{fig:english}
\end{figure}

\begin{figure}
    \centering
    \includegraphics[width=0.95\linewidth]{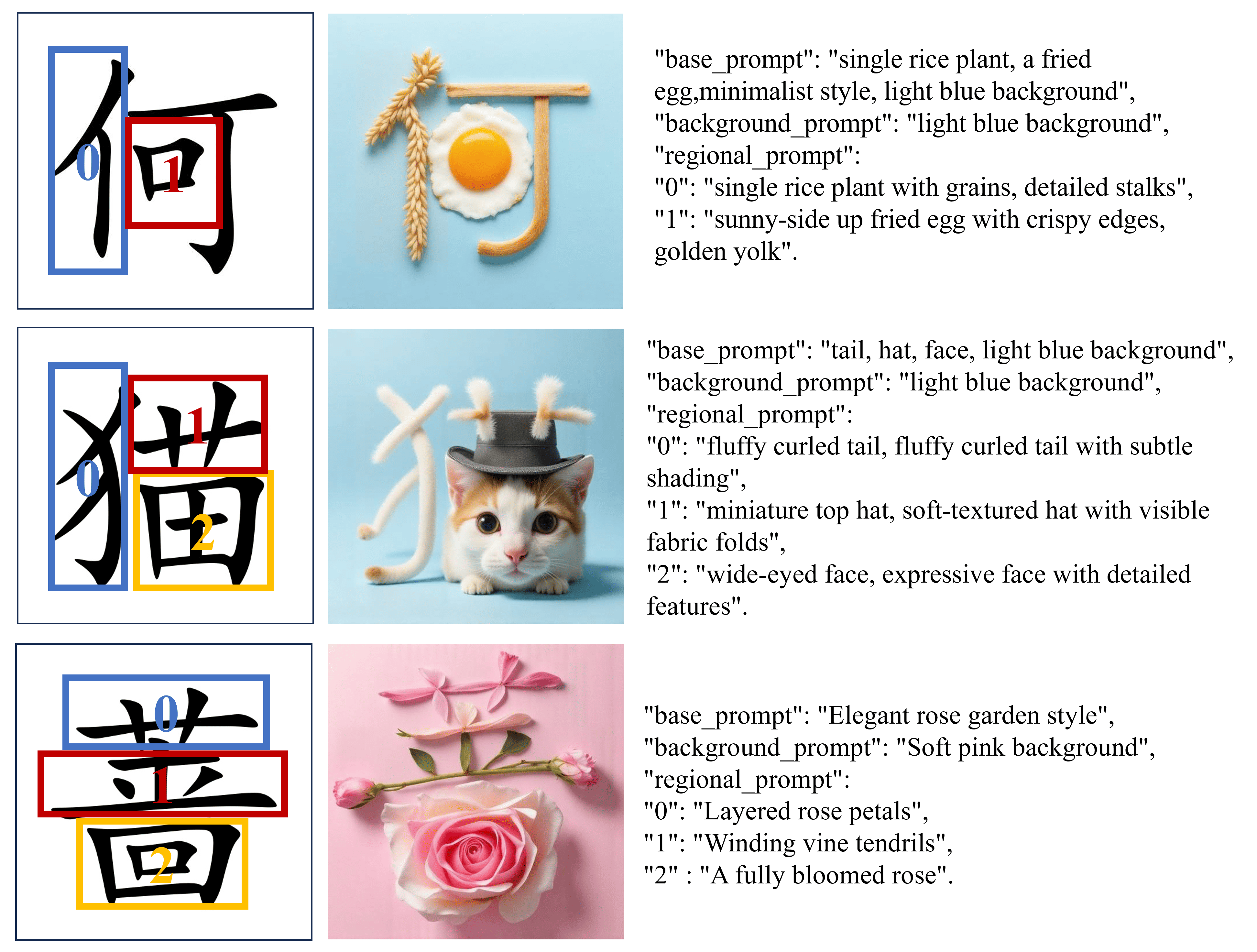}
    \caption{\textbf{Multi-Regional Generation Results.} Our method supports multi-regional generation by drawing various masks.}
    \label{fig:multiregion}
\end{figure}

\begin{figure}
    \centering
    \includegraphics[width=0.95\linewidth]{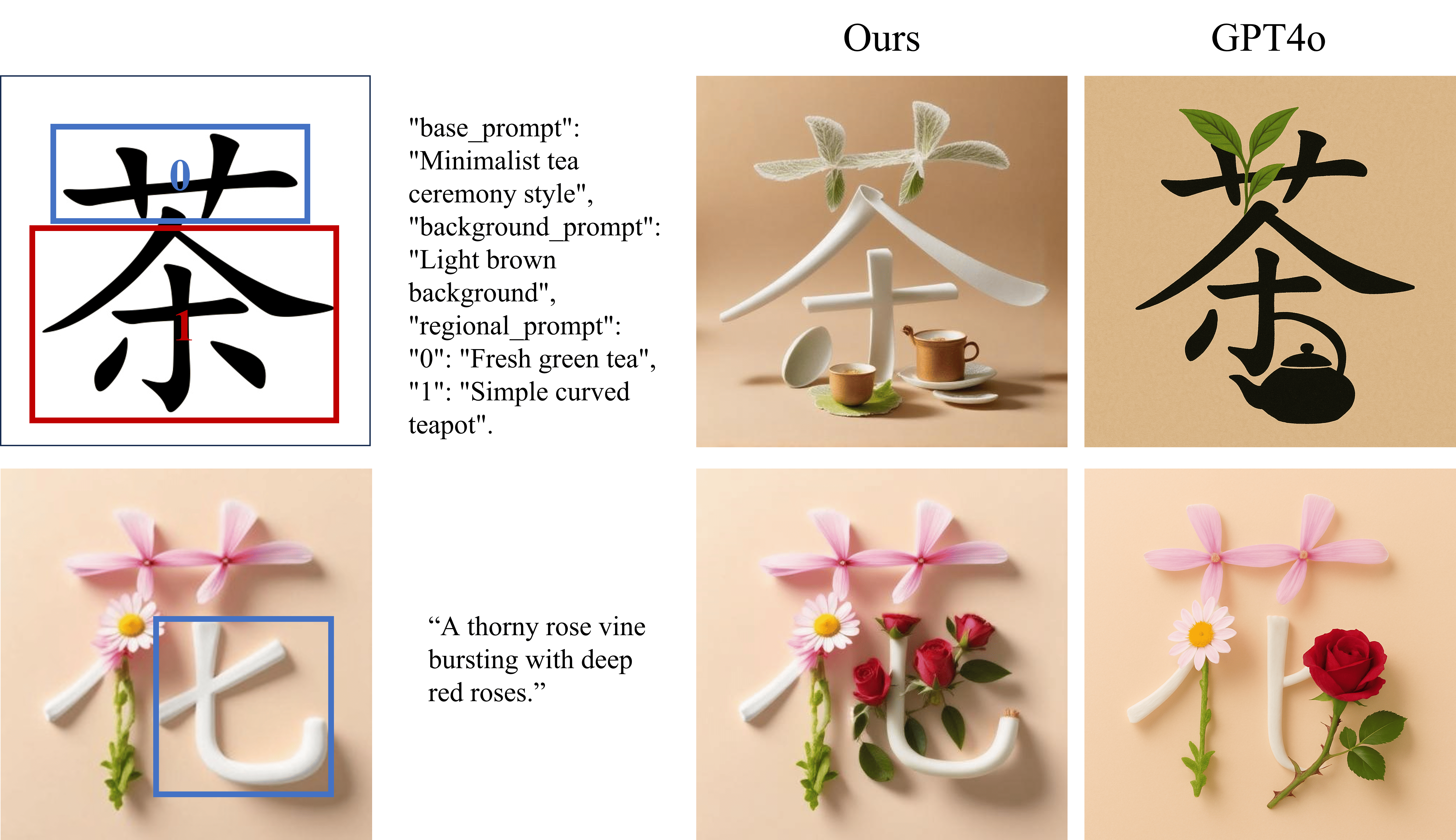}
    \caption{\textbf{Comparison.} We compare our method with \textit{GPT4o}, demonstrating superior semantic consistency in multi-regional generation and better preservation of unedited regions during continuous editing.}
    \label{fig:cmp2}
\end{figure}

\subsection{Implementation Details}

We collected a dataset comprising 176 Chinese words and 179 English words (including single- and multi-character terms) sourced from publicly available online resources. The dataset was constructed through a systematic three-step process: Text Generation: First, we utilized ChatGPT to generate detailed textual descriptions for word stylization. Image Retrieval and Filtering: These descriptions were then used to query the Google web search engine, retrieving relevant visual references. The resulting images were rigorously filtered to ensure quality and relevance. Refinement: Finally, the filtered visual data was fed back into ChatGPT to iteratively refine and enhance the original textual descriptions, ensuring alignment between linguistic and visual representations.

To quantitatively evaluate the generated results, we use CLIP~\cite{radford2021learning} image-text similarity and FID \cite{heusel2017gans} scores to assess the overall quality and coherence. 
Additionally, we have conducted a user study to further assess the effectiveness of our method.

We adopt FLUX.1 \cite{flux2024} as the baseline diffusion model and fine-tune it on our dataset using OminiControl \cite{tan2024ominicontrol}, which incorporates depth maps as conditional inputs. In OminiControl, the depth map is first encoded into the latent space via the pretrained VAE of the diffusion model and then concatenated with noisy image tokens to form a unified input sequence. This sequence is processed using multi-modality attention (MM-Attention) \cite{wei2020multi,esser2024scaling}, enabling joint modeling of multimodal inputs for controllable generation.
During fine-tuning, only the Transformer backbone is updated using LoRA adapters, while the text encoder and VAE modules remain frozen. Depth maps are extracted using a pretrained depth estimation model \cite{depthanything} and serve as conditioning inputs throughout training. We optimize the model using the Prodigy optimizer \cite{mishchenko2024prodigy} with a learning rate of 1.0. Training is conducted on an NVIDIA RTX 5880 GPU with 48 GB RAM, and gradient checkpointing is enabled for memory efficiency. The model is fine-tuned for 80,000 steps.

\subsection{Comparisons}

\begin{table}[t]
\centering
\scalebox{1.0}{
\begin{tabular}{cccc}
\toprule
Method & MetaDesigner& VitaGlyph & Ours \\ \hline
CLIP-Score &  26.33&  25.29&  \textbf{27.52}\\
FID &  181.88&  203.43&   \textbf{141.45}\\
\bottomrule
\end{tabular}}
\caption{\textbf{Quantitative Comparison.} We provide quantitative results compared with \textit{MetaDesigner} and \textit{VitaGlyph}.}
\label{tab:quan}
\end{table}

We compare our method with several recent artistic text synthesis approaches, including \textit{VitaGlyph} \cite{feng2024vitaglyph} and \textit{MetaDesigner} \cite{hemetadesigner}. Since \textit{MetaDesigner} has not released its source code, we conduct comparisons using its online demo. 
As \textit{VitaGlyph} supports only two-region generation (the subject and the surrounding area), we conduct the comparison using two regional masks, even though our method supports multi-regional generation, as demonstrated in Fig.~\ref{fig:multiregion}.
As \textit{MetaDesigner} does not natively support local stylization, we adapt it by introducing localized descriptions to enable a fair comparison in this aspect, as shown in Fig.~\ref{fig:cmp}.

\begin{figure*}
    \centering
    \includegraphics[width=0.95\linewidth]{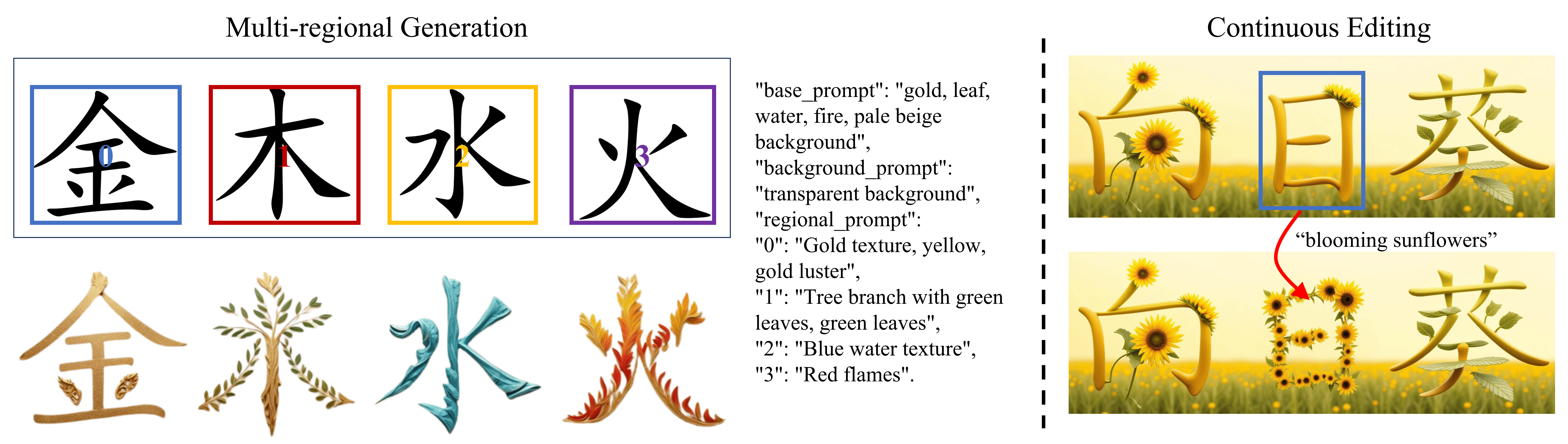}
    \caption{\textbf{Multi-regional Generation and Continuous Editing for Multi-character.} Both our multi-regional attention and noise blending techniques effectively support multi-character scenarios.}
    \label{fig:multi-char}
\end{figure*}

\begin{figure}
    \centering
    \includegraphics[width=0.90\linewidth]{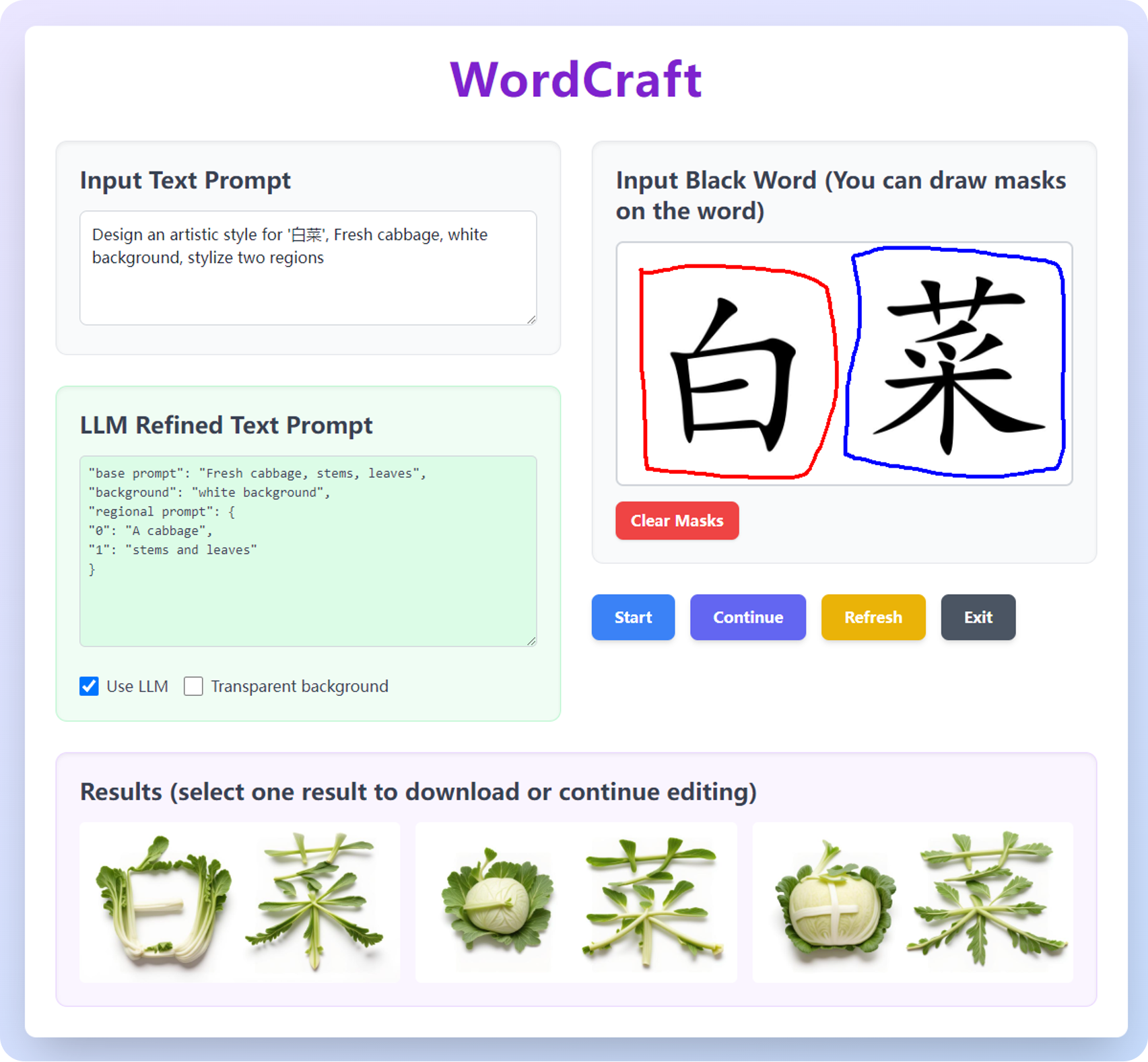}
    \caption{\textbf{User Interface.} We design a user interface for our method.}
    \label{fig:ui}
\end{figure}

\begin{figure*}
    \centering
    \includegraphics[width=0.90\linewidth]{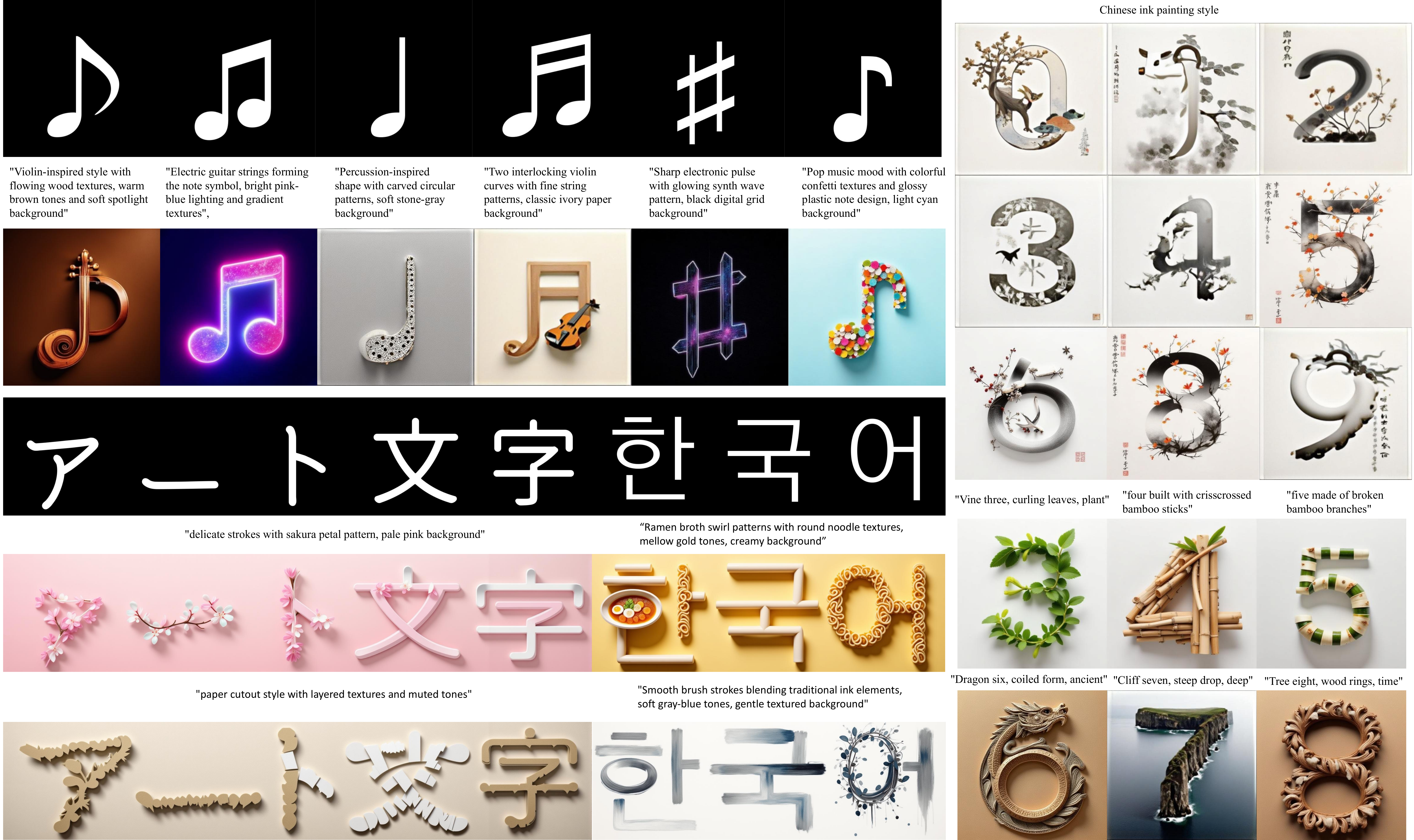}
    \caption{\textbf{Global Generation. }Generation on music notations, Arabic numerals, Japanese and Korean characters.}
    \label{fig:glb1}
\end{figure*}

\begin{figure*}
    \centering
    \includegraphics[width=0.90\linewidth]{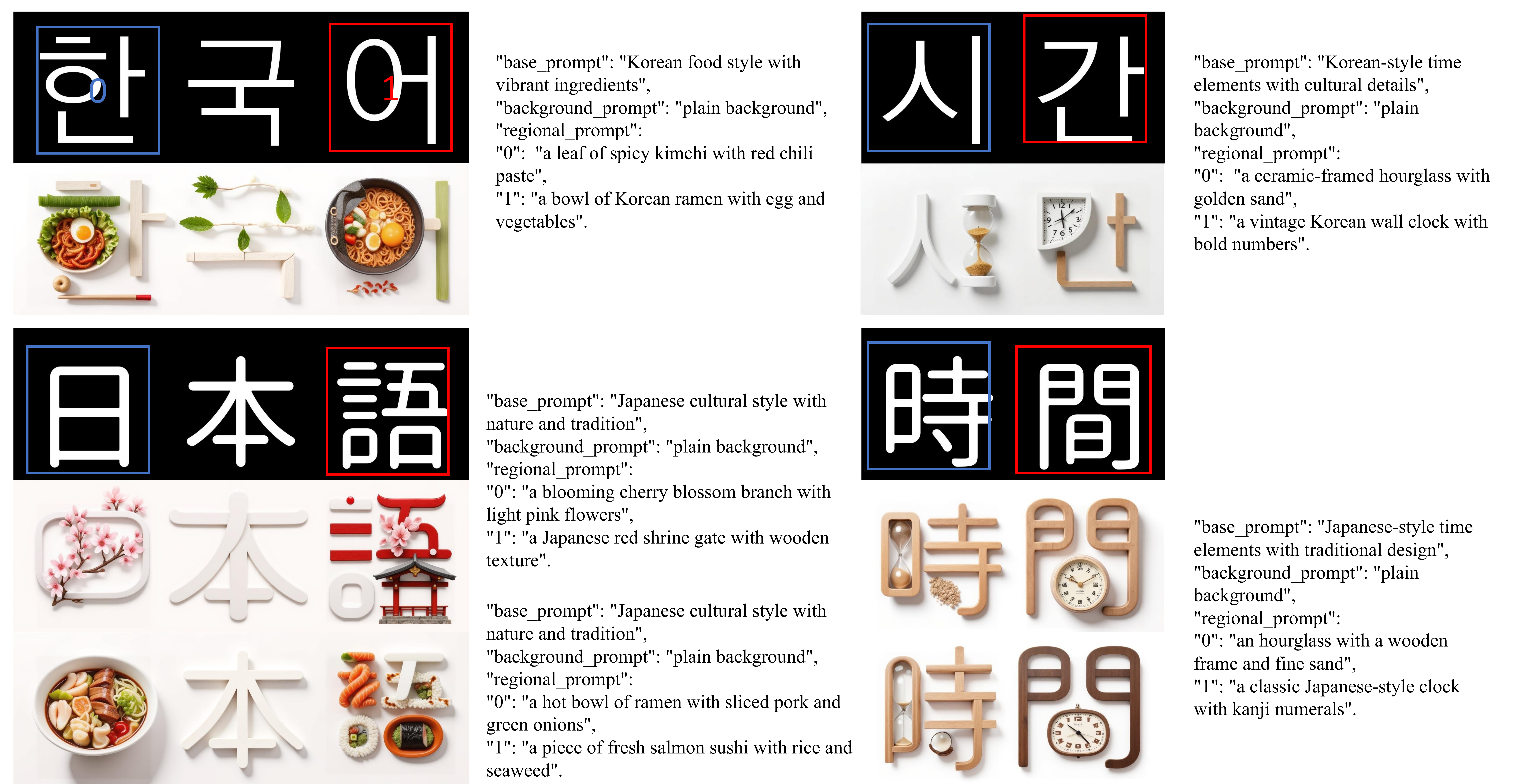}
    \caption{\textbf{Diverse Language Support. }We show \textit{WordCraft}'s capability for Korean and Japanese character generation.}
    \label{fig:kr_jp}
\end{figure*}

\begin{figure*}
    \centering
    \includegraphics[width=0.95\linewidth]{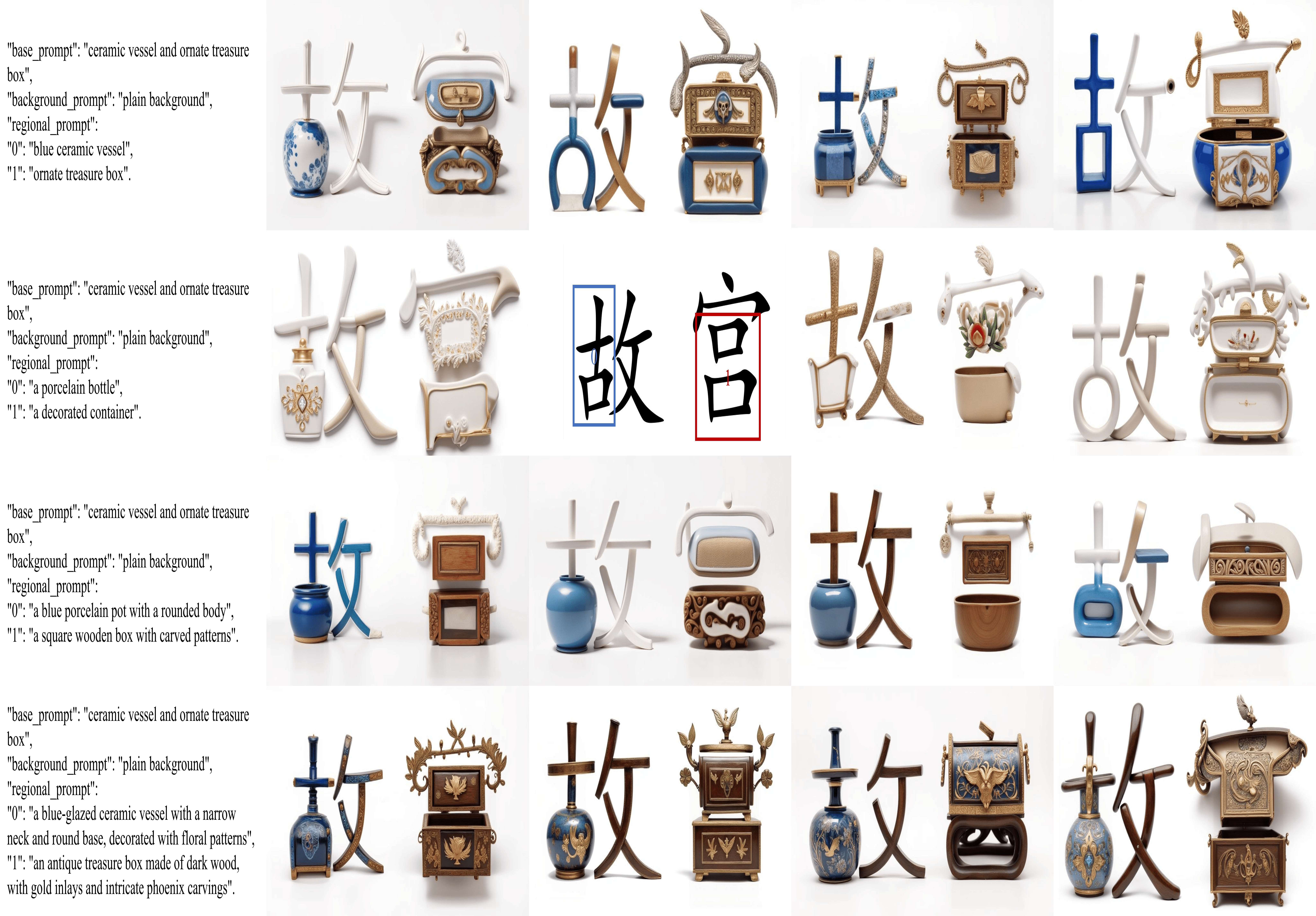}
    \caption{\textbf{Results with Diverse Text Prompts.} Our method supports diverse text descriptions and for the same text prompt, we can generate various results.}
    \label{fig:diverse}
\end{figure*}

\begin{figure*}
    \centering
    \includegraphics[width=0.82\linewidth]{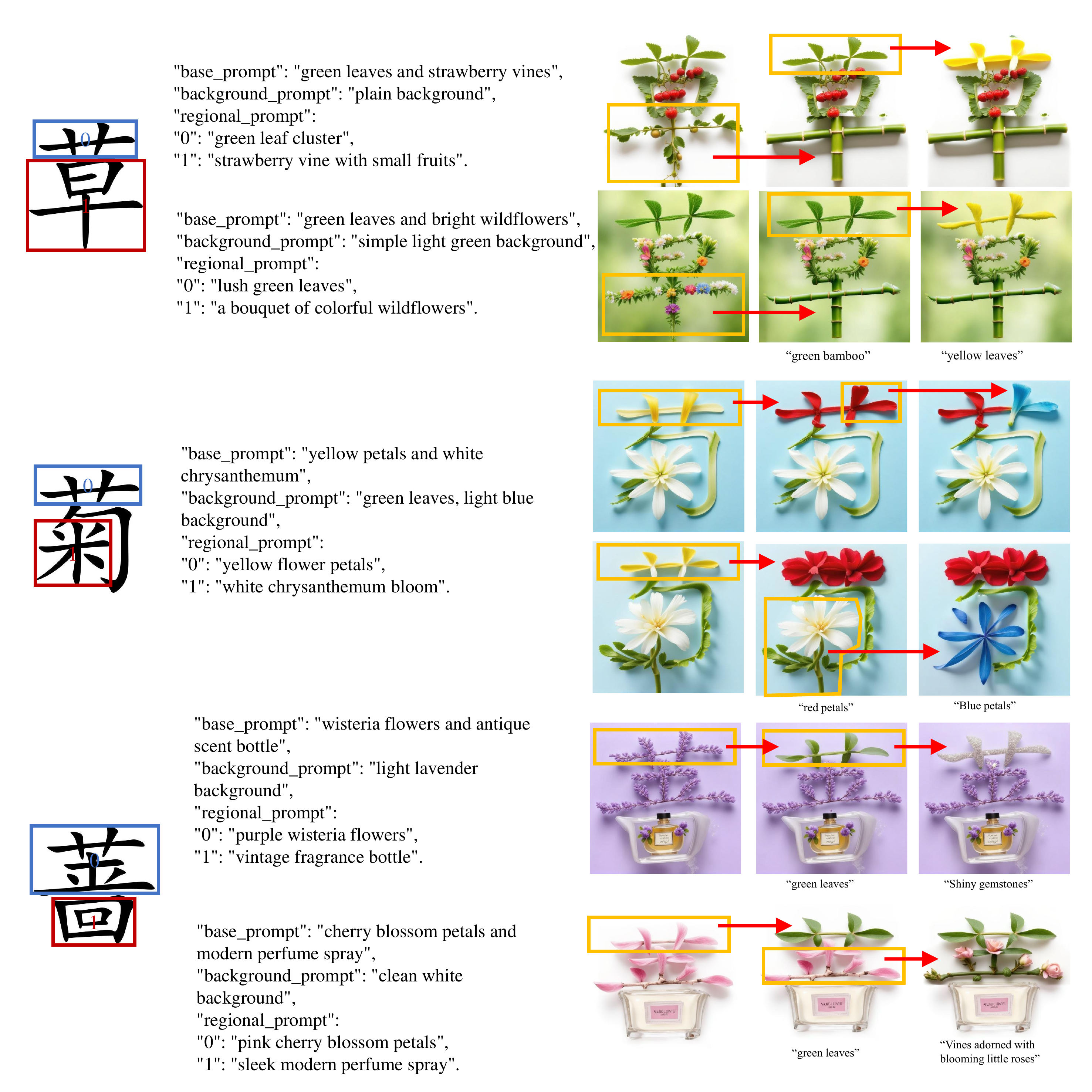}
    \caption{\textbf{Interactive Support.} Our method support long term user continuous editing.}
    \label{fig:con}
\end{figure*}

\noindent\textbf{Quantitative Comparison.} We use 2,000 regional generated samples conditioned on diverse text prompts for quantitative evaluation. As shown in Table~\ref{tab:quan}, our method achieves superior performance across key metrics. Specifically, it surpasses \textit{VitaGlyph} by a notable margin in CLIP score (+2.23\%), indicating better alignment between the generated images and the input text prompts. Furthermore, it achieves a significantly lower FID score compared to \textit{MetaDesigner} (141.45 vs. 181.88), reflecting higher visual realism and fidelity.
These improvements underscore the effectiveness of our multi-regional control mechanism and our model’s ability to maintain semantic coherence across complex, localized edits. The results suggest that our method not only generates more text-consistent images but also maintains visual quality at a level superior to existing state-of-the-art approaches.

\noindent\textbf{Qualitative Comparison.} We present qualitative comparisons in Fig.~\ref{fig:cmp}, where our method consistently outperforms \textit{VitaGlyph} and \textit{MetaDesigner} in terms of visual quality and semantic coherence in regional stylization. The artistic words generated by our approach exhibit more accurate and localized semantic stylization, demonstrating finer control over region-specific edits.
In contrast, \textit{VitaGlyph} and \textit{MetaDesigner} often produce noticeable artifacts and semantic inconsistencies, particularly when handling complex, multi-character terms or localized modifications. For example, when attempting to stylize a character with the description ``five large white petals arranged radially, flat and facing front,'' both baselines fail to produce coherent results that align with the text. Similarly, they struggle to maintain geometry structure when generating `porcelain' or `lotus', often generating overly delicate structures.
Furthermore, our method excels at handling background modifications conditioned on background-specific text prompts. For instance, it successfully generates a pure white background when instructed, while \textit{VitaGlyph} and \textit{MetaDesigner} tend to blend background elements with foreground colors, failing to isolate regions effectively.
Another key advantage of our approach is continued editing: users can iteratively refine and modify unsatisfactory regions until the desired outcome is achieved. This capability is not supported by either \textit{VitaGlyph} or \textit{MetaDesigner}, limiting their flexibility in interactive or user-in-the-loop editing workflows.

We also compare our method with the recent multimodal understanding and generation work, \textit{GPT4o}, as shown in Fig.~\ref{fig:cmp2}. While \textit{GPT4o} excels at generating high-quality artist textures, our method outperforms it in terms of semantic consistency in multi-regional generation and better preservation of unedited regions during continuous editing.

\begin{table}[]
\scalebox{0.9}{
\begin{tabular}{ccccc}
\toprule
Method & Aesthetic & Text Align. & Legibility & ACC \\ \hline
MetaDesigner & 3.01 & 2.98 & 4.12 & 3.37\\
VitaGlyph & 2.68 & 2.89 & 2.97 & 2.84 \\
Ours & \textbf{3.75} & \textbf{4.12} & \textbf{4.32 }& \textbf{4.06}\\
\bottomrule
\end{tabular}}
\caption{\textbf{User study.} Our method outperforms the comparisons in three aspects: Aesthetic, Text Alignment, and Character Legibility.}
\vspace{-8mm}
\label{tab:us}
\end{table}

To further evaluate the effectiveness of our method, we conducted a user study focusing on three key aspects: overall aesthetic appeal, relevance to the textual input, and consistency of local edits. Participants were shown images generated by our method alongside outputs from several baselines, including \textit{MetaDesigner} and \textit{VitaGlyph}, for both single-character and multi-character text cases. For each example, users were asked to score three criteria (Aesthetic, Text Alignment, and Character Legibility) on a scale from 0 to 5, with 5 being the best.
We collected 30 valid responses for 10 questions and calculated the average score for each aspect. As summarized in Table~\ref{tab:us}, the results demonstrate that our method consistently outperforms existing approaches in terms of Aesthetic, Text Alignment, and Character Legibility, confirming its effectiveness from a human evaluation perspective.

\subsection{Ablation Study}

To assess each technical component in our proposed method, we conduct an ablation study by systematically removing key modules and measuring their impact on overall performance.

\noindent\textbf{w/o LLM.} In the absence of the LLM component, the method encounters significant limitations in interpreting user-provided prompts that are arbitrary, unconstrained, and often complex. For instance, in multi-regional generation tasks, users must manually provide structured and explicitly segmented text descriptions. In contrast, with the LLM engine, users can simply input a natural, free-form sentence without worrying about structural constraints, resulting in a more user-friendly and intuitive experience. 
Moreover, our LLM component is capable of converting abstract concepts into detailed and concrete descriptions, which helps generate more intricate details. As illustrated in Fig.~\ref{fig:albation}, compared to the method \textit{w/o LLM}, our approach effectively interprets abstract input and generates more precise, context-aware prompts, resulting in improved outcomes. However, users can still choose to use abstract descriptions for generation, catering to various generation intentions.

\noindent\textbf{w/o Regional Attention.} Without the regional attention mechanism, a straightforward approach to achieve multi-regional stylization is to describe all regional changes within a single text prompt, as illustrated in the method \textit{w/o RA} in Fig.~\ref{fig:albation}. However, relying solely on such a single prompt often leads to suboptimal results—geometry and texture generation in each region fail to accurately reflect the intended descriptions. This is because the diffusion model alone lacks the capacity to effectively disentangle and represent the diverse semantics embedded in a compound prompt. For example, in the second case in Fig.~\ref{fig:albation}, the generation of ``Entwined delicate vines'' is ambiguous, and the background color is not correctly generated.
In contrast, our regional attention explicitly captures the relationship between each region's text prompt and its corresponding visual effect, leading to more accurate and visually coherent results.

\noindent\textbf{w/o Noise Blending.} Our noise blending technique is designed to support continuous editing by preserving important visual context during modification. 
In the absence of this technique, an intuitive approach is to mask and blend the latents corresponding to the original image and the edited region to generate the target image, as denoted by \textit{w/o NB-LB} in Fig.~\ref{fig:albation}. We find that this method often introduces disharmony around the boundaries, with the style of the edited region not aligning with the rest of the image. All cases in Fig.~\ref{fig:albation} exhibit these issues. Another common alternative is to use inpainting-based methods, such as FLUX-Inpainting, to modify the targeted region.
As shown in \textit{w/o NB-Inp} in  Fig.~\ref{fig:albation}, while such methods can perform the edit, they often produce undesired results, particularly in failing to preserve essential details of the original content—such as the identity and structure of a character. A more intuitive solution might involve local depth-guided inpainting; however, training such a model for artistic image synthesis would require a specialized dataset containing local depth maps, region-specific descriptions, and paired ground truth images—resources that are difficult to obtain. In contrast, our training-free noise blending method enables seamless integration of the edited region by blending its noise with the background noise during the denoising process, effectively preserving contextual consistency without the need for additional data or model retraining.

\subsection{Visual Results}

In Fig.~\ref{fig:multiregion}, we present the results of multi-regional editing. For example, different parts of the word are edited with distinct styles: in the first region, layered rose petals are generated, while in the second region of the last case, a fully bloomed rose appears. Thanks to our multi-regional attention mechanism, the model can effectively localize and apply diverse styles within a single word, enabling fine-grained and coherent artistic typography.

In Fig.~\ref{fig:continuous}, we present the results of continuous editing. Using our noise blending technique, users can iteratively refine existing artistic typography within our framework. As shown, the continuous editing results align well with user intent while preserving the overall structure of the original words to a reasonable extent. In conclusion, our method modifies only the masked region, preserving the other regions unchanged and ensuring a harmonious boundary.

Moreover, our multi-region attention and noise blending techniques support not only single characters but also multiple characters. We show these cases in Fig.~\ref{fig:multi-char}. This demonstrates the extensibility of our techniques.

In Fig.~\ref{fig:english}, we demonstrate that our method supports English characters, enabling multi-regional generation and continuous editing. This capability allows us to seamlessly generate stylized typography for English words while preserving both readability and aesthetic consistency.

\subsection{User Interface}

The \textit{WordCraft} user interface, as shown in Fig.~\ref{fig:ui}, is designed to provide a dynamic and interactive experience for artistic typography creation. Users begin by inputting a text prompt that describes the word they wish to stylize and the target style. This prompt is then refined using a large language model (LLM), which enhances and structures the description for more complex creative outputs. Automatically, the system generates the ``black word'' based on the user's input through our Character Parameterization. Users can then interact with the generated word, drawing and masking specific regions to customize the styling.
The interface includes a ``Results'' section, where users can preview various generated outcomes and select one for further refinement or download. This is made easy with three key buttons: the ``Start'' button begins the process of generating typography, the ``Refresh'' button creates a new set of results for exploration, and the ``Continue'' button allows users to refine a previously generated result. When ``Continue'' is selected, the chosen result is displayed in the top-right corner for further editing.
Additionally, two optional features provide enhanced flexibility. The ``Use LLM'' option converts abstract descriptions into more concrete prompts, making it easier to create specific styles. The ``Transparent Background'' option, when enabled, removes the background from the results, offering a cleaner output for different design needs.
Overall, \textit{WordCraft} provides a structured, intuitive interface that enables users to interactively refine their typography designs, giving them full creative control over the final output.

\subsection{More Results}

\noindent\textbf{More Global Generation Results. }
We provide more global generation results in Fig.~\ref{fig:glb1}. Our method can support globally stylizing music notations, Arabic numerals, Japanese, and Korean characters, even if these characters do not exist in our training dataset.

\noindent\textbf{More Results on Multi-regional Generation. }In Fig.~\ref{fig:kr_jp}, we show our multi-regional generation results on out-of-domain characters. We show \textit{WordCraft}’s multi-regional generation capability for Korean and Japanese character generation.

\noindent\textbf{Continuous Editing Workflow.}
Our method enables a continuous editing workflow for users. Users can iteratively mask the region for continuous editing until the desired results are obtained. We show the results in Fig.~\ref{fig:con}.

\noindent\textbf{Generalization Evaluation.}
We provide an illustration for the generalization ability of our method in handling diverse inputs (i.e., musical notations, numbers, Japanese and Korean characters) in Fig. \ref{fig:diverse}.
out-of-domain evaluation: (no training, direct evaluation)

\noindent\textbf{Variety Evaluation.}
We show diverse text prompts from coarse to fine in Fig.~\ref{fig:diverse}. Our method supports diverse text descriptions, and for the same text prompt, we can generate various results.

\section{Conclusion}

In conclusion, this paper presents \textit{WordCraft}, an innovative interactive system for artistic typography. By integrating a novel regional attention mechanism, \textit{WordCraft} enables precise, multi-region styling that is essential for handling complex, semantically rich characters and multi-character compositions. The system's noise blending technique ensures continuous, iterative refinement, allowing users to edit typography without compromising visual quality. Additionally, the incorporation of a LLM empowers the system to interpret and structure diverse user prompts, providing an intuitive and flexible interface for generating both concrete and abstract visual effects.
In summary, \textit{WordCraft} makes significant strides in artistic typography by enhancing user interactivity and creative control.

\begin{figure}
    \centering
    \includegraphics[width=\linewidth]{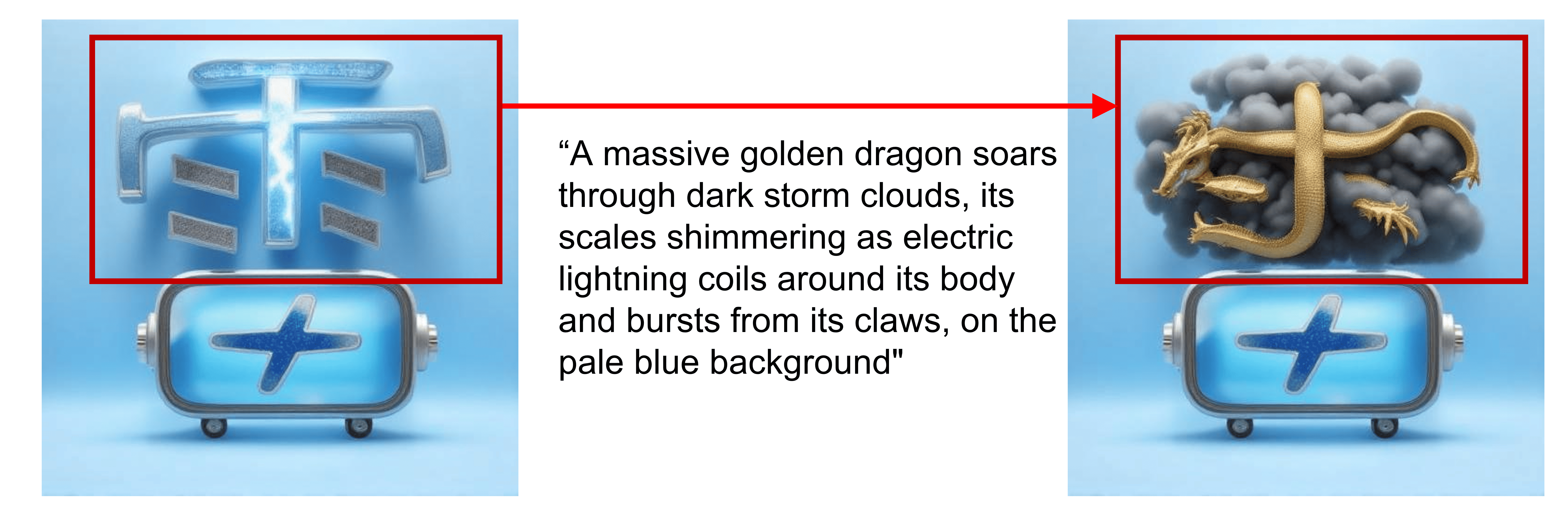}
    \caption{\textbf{Limitation}. Our method may fail when the text description are very fine-grained and detailed, the results may not be consistent with the semantics.}
    \label{fig:limit}
\end{figure}

Despite all the generation results in the main text and the supplemental materials, our method may generate with less satisfactory results. We show such a case in Fig. \ref{fig:limit}: when the prompts are too detailed, the diffusion model may have trouble handling the fine details in the prompt. With the advancement of generation backbone models, this problem can be effectively resolved.

{
    \small
    \bibliographystyle{ieeenat_fullname}
    \bibliography{cite}

\begin{thebibliography}{44}
\providecommand{\natexlab}[1]{#1}
\providecommand{\url}[1]{\texttt{#1}}
\expandafter\ifx\csname urlstyle\endcsname\relax
  \providecommand{\doi}[1]{doi: #1}\else
  \providecommand{\doi}{doi: \begingroup \urlstyle{rm}\Url}\fi

\bibitem[Ahn et~al.(2024)Ahn, Verma, Lou, Liu, Zhang, and Yin]{ahn2024large}
Janice Ahn, Rishu Verma, Renze Lou, Di Liu, Rui Zhang, and Wenpeng Yin.
\newblock Large language models for mathematical reasoning: Progresses and challenges.
\newblock In \emph{18th Conference of the European Chapter of the Association for Computational Linguistics, EACL 2024-Student Research Workshop, SRW 2024}, pages 225--237. Association for Computational Linguistics (ACL), 2024.

\bibitem[Azadi et~al.(2018)Azadi, Fisher, Kim, Wang, Shechtman, and Darrell]{azadi2018multi}
Samaneh Azadi, Matthew Fisher, Vladimir~G Kim, Zhaowen Wang, Eli Shechtman, and Trevor Darrell.
\newblock Multi-content gan for few-shot font style transfer.
\newblock In \emph{Proceedings of the IEEE conference on computer vision and pattern recognition}, pages 7564--7573, 2018.

\bibitem[Bhat et~al.(2024)Bhat, Mitra, and Wonka]{bhat2024loosecontrol}
Shariq~Farooq Bhat, Niloy Mitra, and Peter Wonka.
\newblock Loosecontrol: Lifting controlnet for generalized depth conditioning.
\newblock In \emph{ACM SIGGRAPH 2024 Conference Papers}, pages 1--11, 2024.

\bibitem[Esser et~al.(2024)Esser, Kulal, Blattmann, Entezari, M{\"u}ller, Saini, Levi, Lorenz, Sauer, Boesel, et~al.]{esser2024scaling}
Patrick Esser, Sumith Kulal, Andreas Blattmann, Rahim Entezari, Jonas M{\"u}ller, Harry Saini, Yam Levi, Dominik Lorenz, Axel Sauer, Frederic Boesel, et~al.
\newblock Scaling rectified flow transformers for high-resolution image synthesis.
\newblock In \emph{Forty-first international conference on machine learning}, 2024.

\bibitem[Feng et~al.(2024)Feng, Zhang, Yu, Ji, Bai, Zhang, and Zuo]{feng2024vitaglyph}
Kailai Feng, Yabo Zhang, Haodong Yu, Zhilong Ji, Jinfeng Bai, Hongzhi Zhang, and Wangmeng Zuo.
\newblock Vitaglyph: Vitalizing artistic typography with flexible dual-branch diffusion models.
\newblock \emph{arXiv preprint arXiv:2410.01738}, 2024.

\bibitem[Fu et~al.(2023)Fu, He, Wang, and Qiao]{fu2023neural}
Bin Fu, Junjun He, Jianjun Wang, and Yu Qiao.
\newblock Neural transformation fields for arbitrary-styled font generation.
\newblock In \emph{Proceedings of the IEEE/CVF conference on computer vision and pattern recognition}, pages 22438--22447, 2023.

\bibitem[He et~al.(2023)He, Cheng, Li, Sun, Xiang, Lin, Kang, Jin, Hu, Luo, et~al.]{he2023wordart}
Jun-Yan He, Zhi-Qi Cheng, Chenyang Li, Jingdong Sun, Wangmeng Xiang, Xianhui Lin, Xiaoyang Kang, Zengke Jin, Yusen Hu, Bin Luo, et~al.
\newblock Wordart designer: User-driven artistic typography synthesis using large language models.
\newblock In \emph{Proceedings of the 2023 Conference on Empirical Methods in Natural Language Processing: Industry Track}, pages 223--232, 2023.

\bibitem[He et~al.(2025)He, Cheng, Li, Sun, He, Xiang, Chen, Lan, Lin, Luo, et~al.]{hemetadesigner}
Jun-Yan He, Zhi-Qi Cheng, Chenyang Li, Jingdong Sun, Qi He, Wangmeng Xiang, Hanyuan Chen, Jin-Peng Lan, Xianhui Lin, Bin Luo, et~al.
\newblock Metadesigner: Advancing artistic typography through ai-driven, user-centric, and multilingual wordart synthesis.
\newblock In \emph{The Thirteenth International Conference on Learning Representations}, 2025.

\bibitem[Heusel et~al.(2017)Heusel, Ramsauer, Unterthiner, Nessler, and Hochreiter]{heusel2017gans}
Martin Heusel, Hubert Ramsauer, Thomas Unterthiner, Bernhard Nessler, and Sepp Hochreiter.
\newblock Gans trained by a two time-scale update rule converge to a local nash equilibrium.
\newblock \emph{Advances in neural information processing systems}, 30, 2017.

\bibitem[Iluz et~al.(2023)Iluz, Vinker, Hertz, Berio, Cohen-Or, and Shamir]{iluz2023word}
Shir Iluz, Yael Vinker, Amir Hertz, Daniel Berio, Daniel Cohen-Or, and Ariel Shamir.
\newblock Word-as-image for semantic typography.
\newblock \emph{ACM Transactions on Graphics (TOG)}, 42\penalty0 (4):\penalty0 1--11, 2023.

\bibitem[Isola et~al.(2017)Isola, Zhu, Zhou, and Efros]{isola2017image}
Phillip Isola, Jun-Yan Zhu, Tinghui Zhou, and Alexei~A Efros.
\newblock Image-to-image translation with conditional adversarial networks.
\newblock In \emph{Proceedings of the IEEE conference on computer vision and pattern recognition}, pages 1125--1134, 2017.

\bibitem[Jing et~al.(2019)Jing, Yang, Feng, Ye, Yu, and Song]{jing2019neural}
Yongcheng Jing, Yezhou Yang, Zunlei Feng, Jingwen Ye, Yizhou Yu, and Mingli Song.
\newblock Neural style transfer: A review.
\newblock \emph{IEEE transactions on visualization and computer graphics}, 26\penalty0 (11):\penalty0 3365--3385, 2019.

\bibitem[Labs(2024)]{flux2024}
Black~Forest Labs.
\newblock Flux.
\newblock \url{https://github.com/black-forest-labs/flux}, 2024.

\bibitem[Li et~al.(2020)Li, Luk{\'a}{\v{c}}, Gharbi, and Ragan-Kelley]{li2020differentiable}
Tzu-Mao Li, Michal Luk{\'a}{\v{c}}, Micha{\"e}l Gharbi, and Jonathan Ragan-Kelley.
\newblock Differentiable vector graphics rasterization for editing and learning.
\newblock \emph{ACM Transactions on Graphics (TOG)}, 39\penalty0 (6):\penalty0 1--15, 2020.

\bibitem[Liu et~al.(2024)Liu, Wei, Liu, Lin, Ren, Xie, and Zuo]{liu2024smartcontrol}
Xiaoyu Liu, Yuxiang Wei, Ming Liu, Xianhui Lin, Peiran Ren, Xuansong Xie, and Wangmeng Zuo.
\newblock Smartcontrol: Enhancing controlnet for handling rough visual conditions.
\newblock In \emph{European Conference on Computer Vision}, pages 1--17. Springer, 2024.

\bibitem[Mao et~al.(2022)Mao, Yang, Shi, Liu, and Wang]{mao2022intelligent}
Wendong Mao, Shuai Yang, Huihong Shi, Jiaying Liu, and Zhongfeng Wang.
\newblock Intelligent typography: Artistic text style transfer for complex texture and structure.
\newblock \emph{IEEE Transactions on Multimedia}, 25:\penalty0 6485--6498, 2022.

\bibitem[Mishchenko and Defazio(2024)]{mishchenko2024prodigy}
Konstantin Mishchenko and Aaron Defazio.
\newblock Prodigy: An expeditiously adaptive parameter-free learner.
\newblock In \emph{Forty-first International Conference on Machine Learning}, 2024.

\bibitem[Mu et~al.(2024)Mu, Chen, Chen, Gu, Bao, Chen, Li, and Yuan]{mu2024fontstudio}
Xinzhi Mu, Li Chen, Bohan Chen, Shuyang Gu, Jianmin Bao, Dong Chen, Ji Li, and Yuhui Yuan.
\newblock Fontstudio: shape-adaptive diffusion model for coherent and consistent font effect generation.
\newblock In \emph{European Conference on Computer Vision}, pages 305--322. Springer, 2024.

\bibitem[Nam et~al.(2024)Nam, Macvean, Hellendoorn, Vasilescu, and Myers]{nam2024using}
Daye Nam, Andrew Macvean, Vincent Hellendoorn, Bogdan Vasilescu, and Brad Myers.
\newblock Using an llm to help with code understanding.
\newblock In \emph{Proceedings of the IEEE/ACM 46th International Conference on Software Engineering}, pages 1--13, 2024.

\bibitem[Pan et~al.(2020)Pan, Luo, Yang, and Li]{pan2020multi}
Zexu Pan, Zhaojie Luo, Jichen Yang, and Haizhou Li.
\newblock Multi-modal attention for speech emotion recognition.
\newblock \emph{arXiv preprint arXiv:2009.04107}, 2020.

\bibitem[Podell et~al.(2023)Podell, English, Lacey, Blattmann, Dockhorn, M{\"u}ller, Penna, and Rombach]{podell2023sdxl}
Dustin Podell, Zion English, Kyle Lacey, Andreas Blattmann, Tim Dockhorn, Jonas M{\"u}ller, Joe Penna, and Robin Rombach.
\newblock Sdxl: Improving latent diffusion models for high-resolution image synthesis.
\newblock \emph{arXiv preprint arXiv:2307.01952}, 2023.

\bibitem[Radford et~al.(2021)Radford, Kim, Hallacy, Ramesh, Goh, Agarwal, Sastry, Askell, Mishkin, Clark, et~al.]{radford2021learning}
Alec Radford, Jong~Wook Kim, Chris Hallacy, Aditya Ramesh, Gabriel Goh, Sandhini Agarwal, Girish Sastry, Amanda Askell, Pamela Mishkin, Jack Clark, et~al.
\newblock Learning transferable visual models from natural language supervision.
\newblock In \emph{International conference on machine learning}, pages 8748--8763. PmLR, 2021.

\bibitem[Ramesh et~al.(2022)Ramesh, Dhariwal, Nichol, Chu, and Chen]{ramesh2022hierarchical}
Aditya Ramesh, Prafulla Dhariwal, Alex Nichol, Casey Chu, and Mark Chen.
\newblock Hierarchical text-conditional image generation with clip latents.
\newblock \emph{arXiv preprint arXiv:2204.06125}, 1\penalty0 (2):\penalty0 3, 2022.

\bibitem[Rombach et~al.(2022)Rombach, Blattmann, Lorenz, Esser, and Ommer]{rombach2022high}
Robin Rombach, Andreas Blattmann, Dominik Lorenz, Patrick Esser, and Bj{\"o}rn Ommer.
\newblock High-resolution image synthesis with latent diffusion models.
\newblock In \emph{Proceedings of the IEEE/CVF conference on computer vision and pattern recognition}, pages 10684--10695, 2022.

\bibitem[Saharia et~al.(2022)Saharia, Chan, Saxena, Li, Whang, Denton, Ghasemipour, Gontijo~Lopes, Karagol~Ayan, Salimans, et~al.]{saharia2022photorealistic}
Chitwan Saharia, William Chan, Saurabh Saxena, Lala Li, Jay Whang, Emily~L Denton, Kamyar Ghasemipour, Raphael Gontijo~Lopes, Burcu Karagol~Ayan, Tim Salimans, et~al.
\newblock Photorealistic text-to-image diffusion models with deep language understanding.
\newblock \emph{Advances in neural information processing systems}, 35:\penalty0 36479--36494, 2022.

\bibitem[Tan et~al.(2024)Tan, Liu, Yang, Xue, and Wang]{tan2024ominicontrol}
Zhenxiong Tan, Songhua Liu, Xingyi Yang, Qiaochu Xue, and Xinchao Wang.
\newblock Ominicontrol: Minimal and universal control for diffusion transformer.
\newblock \emph{arXiv preprint arXiv:2411.15098}, 3, 2024.

\bibitem[Tanveer et~al.(2023)Tanveer, Wang, Mahdavi-Amiri, and Zhang]{tanveer2023ds}
Maham Tanveer, Yizhi Wang, Ali Mahdavi-Amiri, and Hao Zhang.
\newblock Ds-fusion: Artistic typography via discriminated and stylized diffusion.
\newblock In \emph{Proceedings of the IEEE/CVF International Conference on Computer Vision}, pages 374--384, 2023.

\bibitem[Turner et~al.(1996)Turner, Wilhelm, and Lemberg]{turner1996freetype}
David Turner, Robert Wilhelm, and Werner Lemberg.
\newblock Freetype, 1996.

\bibitem[Wang et~al.(2023)Wang, Wu, Liu, Li, Meng, and Meng]{wang2023anything}
Changshuo Wang, Lei Wu, Xiaole Liu, Xiang Li, Lei Meng, and Xiangxu Meng.
\newblock Anything to glyph: artistic font synthesis via text-to-image diffusion model.
\newblock In \emph{SIGGRAPH Asia 2023 Conference Papers}, pages 1--11, 2023.

\bibitem[Wang et~al.(2024)Wang, Zhong, Chai, He, Chen, and Liao]{wang2024chat2layout}
Can Wang, Hongliang Zhong, Menglei Chai, Mingming He, Dongdong Chen, and Jing Liao.
\newblock Chat2layout: Interactive 3d furniture layout with a multimodal llm.
\newblock \emph{arXiv preprint arXiv:2407.21333}, 2024.

\bibitem[Wang et~al.(2019)Wang, Liu, Yang, and Guo]{wang2019typography}
Wenjing Wang, Jiaying Liu, Shuai Yang, and Zongming Guo.
\newblock Typography with decor: Intelligent text style transfer.
\newblock In \emph{Proceedings of the IEEE/CVF Conference on Computer Vision and Pattern Recognition}, pages 5889--5897, 2019.

\bibitem[Wei et~al.(2020)Wei, Zhang, Li, Zhang, and Wu]{wei2020multi}
Xi Wei, Tianzhu Zhang, Yan Li, Yongdong Zhang, and Feng Wu.
\newblock Multi-modality cross attention network for image and sentence matching.
\newblock In \emph{Proceedings of the IEEE/CVF conference on computer vision and pattern recognition}, pages 10941--10950, 2020.

\bibitem[Yang et~al.(2024)Yang, Kang, Huang, Xu, Feng, and Zhao]{depthanything}
Lihe Yang, Bingyi Kang, Zilong Huang, Xiaogang Xu, Jiashi Feng, and Hengshuang Zhao.
\newblock Depth anything: Unleashing the power of large-scale unlabeled data.
\newblock In \emph{CVPR}, 2024.

\bibitem[Yang et~al.(2019{\natexlab{a}})Yang, Liu, Wang, and Guo]{yang2019tet}
Shuai Yang, Jiaying Liu, Wenjing Wang, and Zongming Guo.
\newblock Tet-gan: Text effects transfer via stylization and destylization.
\newblock In \emph{Proceedings of the AAAI Conference on Artificial Intelligence}, pages 1238--1245, 2019{\natexlab{a}}.

\bibitem[Yang et~al.(2019{\natexlab{b}})Yang, Wang, Wang, Xu, Liu, and Guo]{yang2019controllable}
Shuai Yang, Zhangyang Wang, Zhaowen Wang, Ning Xu, Jiaying Liu, and Zongming Guo.
\newblock Controllable artistic text style transfer via shape-matching gan.
\newblock In \emph{Proceedings of the IEEE/CVF International Conference on Computer Vision}, pages 4442--4451, 2019{\natexlab{b}}.

\bibitem[Yang et~al.(2021)Yang, Wang, and Liu]{yang2021shape}
Shuai Yang, Zhangyang Wang, and Jiaying Liu.
\newblock Shape-matching gan++: Scale controllable dynamic artistic text style transfer.
\newblock \emph{IEEE Transactions on Pattern Analysis and Machine Intelligence}, 44\penalty0 (7):\penalty0 3807--3820, 2021.

\bibitem[Yang et~al.(2023)Yang, Gui, Yuan, Liang, Ding, Hu, and Chen]{yang2023glyphcontrol}
Yukang Yang, Dongnan Gui, Yuhui Yuan, Weicong Liang, Haisong Ding, Han Hu, and Kai Chen.
\newblock Glyphcontrol: glyph conditional control for visual text generation.
\newblock \emph{Advances in Neural Information Processing Systems}, 36:\penalty0 44050--44066, 2023.

\bibitem[Yao et~al.(2024)Yao, Duan, Xu, Cai, Sun, and Zhang]{yao2024survey}
Yifan Yao, Jinhao Duan, Kaidi Xu, Yuanfang Cai, Zhibo Sun, and Yue Zhang.
\newblock A survey on large language model (llm) security and privacy: The good, the bad, and the ugly.
\newblock \emph{High-Confidence Computing}, page 100211, 2024.

\bibitem[Zhang et~al.(2024)Zhang, Li, Zhang, Cao, Shan, and Liao]{zhang2024humanref}
Jingbo Zhang, Xiaoyu Li, Qi Zhang, Yanpei Cao, Ying Shan, and Jing Liao.
\newblock Humanref: Single image to 3d human generation via reference-guided diffusion.
\newblock In \emph{Proceedings of the IEEE/CVF Conference on Computer Vision and Pattern Recognition}, pages 1844--1854, 2024.

\bibitem[Zhang et~al.(2023)Zhang, Rao, and Agrawala]{zhang2023adding}
Lvmin Zhang, Anyi Rao, and Maneesh Agrawala.
\newblock Adding conditional control to text-to-image diffusion models.
\newblock In \emph{Proceedings of the IEEE/CVF international conference on computer vision}, pages 3836--3847, 2023.

\bibitem[Zhao et~al.(2023)Zhao, Chen, Chen, Bao, Hao, Yuan, and Wong]{zhao2023uni}
Shihao Zhao, Dongdong Chen, Yen-Chun Chen, Jianmin Bao, Shaozhe Hao, Lu Yuan, and Kwan-Yee~K Wong.
\newblock Uni-controlnet: All-in-one control to text-to-image diffusion models.
\newblock \emph{Advances in Neural Information Processing Systems}, 36:\penalty0 11127--11150, 2023.

\bibitem[Zhao et~al.(2025)Zhao, Peng, Yang, Luo, Li, Chen, Yang, He, Zhao, Lu, et~al.]{zhao2025local}
Yibo Zhao, Liang Peng, Yang Yang, Zekai Luo, Hengjia Li, Yao Chen, Zheng Yang, Xiaofei He, Wei Zhao, Qinglin Lu, et~al.
\newblock Local conditional controlling for text-to-image diffusion models.
\newblock In \emph{Proceedings of the AAAI Conference on Artificial Intelligence}, pages 10492--10500, 2025.

\bibitem[Zheng et~al.(2023)Zheng, Chiang, Sheng, Zhuang, Wu, Zhuang, Lin, Li, Li, Xing, et~al.]{zheng2023judging}
Lianmin Zheng, Wei-Lin Chiang, Ying Sheng, Siyuan Zhuang, Zhanghao Wu, Yonghao Zhuang, Zi Lin, Zhuohan Li, Dacheng Li, Eric Xing, et~al.
\newblock Judging llm-as-a-judge with mt-bench and chatbot arena.
\newblock \emph{Advances in Neural Information Processing Systems}, 36:\penalty0 46595--46623, 2023.

\bibitem[Zhu et~al.(2017)Zhu, Park, Isola, and Efros]{zhu2017unpaired}
Jun-Yan Zhu, Taesung Park, Phillip Isola, and Alexei~A Efros.
\newblock Unpaired image-to-image translation using cycle-consistent adversarial networks.
\newblock In \emph{Proceedings of the IEEE international conference on computer vision}, pages 2223--2232, 2017.

\end{thebibliography}
}

\end{document}